%% file: main.tex
\documentclass[final,3p,times,twocolumn,nopreprintline]{elsarticle}  

\usepackage{amssymb}
\usepackage{amsmath}

\usepackage{fancyhdr}
\fancypagestyle{firstpage}{
  \fancyhf{}
  \fancyhead[L]{\textit{Accepted for publication in IEEE Access}}
}

\input{our_paper/defs}

\begin{document}

\begin{frontmatter}

\title{Semantic-aware Random Convolution and Source Matching\\ for Domain Generalization in Medical Image Segmentation}

\author[imi,biophysics,ivc]{Franz~Thaler}
\author[imi,biotechmed]{Martin~Urschler\corref{cor1}}
\cortext[cor1]{Corresponding author: martin.urschler@medunigraz.at}
\author[imi]{Mateusz~Kozi\'{n}ski}
\author[biophysics]{Matthias~AF~Gsell}
\author[biophysics,biotechmed]{Gernot~Plank}
\author[biophysics,avl]{Darko~\v{S}tern}

\affiliation[imi]{organization={Institute for Medical Informatics, Statistics and Documentation, Medical University of Graz}, 
            addressline={Auenbruggerplatz 2}, 
            city={Graz},
            postcode={8010}, 
            country={Austria}}
\affiliation[biophysics]{organization={Gottfried Schatz Research Center: Medical Physics and Biophysics, Medical University of Graz}, 
            addressline={Neue Stiftingtalstraße 6}, 
            city={Graz},
            postcode={8010}, 
            country={Austria}}
\affiliation[ivc]{organization={Institute of Visual Computing, Graz University of Technology}, 
            addressline={Inffeldgasse 16}, 
            city={Graz},
            postcode={8010}, 
            country={Austria}}
\affiliation[avl]{organization={AVL List GmbH}, 
            addressline={Hans-List-Platz 1}, 
            city={Graz},
            postcode={8020}, 
            country={Austria}}
\affiliation[biotechmed]{organization={BioTechMed-Graz}, 
            addressline={Mozartgasse 12}, 
            city={Graz},
            postcode={8010}, 
            country={Austria}}

\begin{abstract}
We tackle the challenging problem of single-source domain generalization (DG) for medical image segmentation, where we train a network on one domain (e.g., CT) and directly apply it to a different domain (e.g., MR) without adapting the model and without requiring images or annotations from the new domain during training.
Our method diversifies the source domain through semantic-aware random convolution, where different regions of a source image are augmented differently at training-time, based on their annotation labels. 
At test-time, we complement the randomization of the training domain via mapping the intensity of target domain images, making them similar to source domain data.
We perform a comprehensive evaluation on a variety of cross-modality and cross-center generalization settings for abdominal, whole-heart and prostate segmentation, where we outperform previous DG techniques in a vast majority of experiments.
Additionally, we also investigate our method when training on whole-heart CT or MR data and testing on the diastolic and systolic phase of cine MR data captured with different scanner hardware.
Overall, our evaluation shows that our method achieves new state-of-the-art performance in DG for medical image segmentation, even matching the performance of the in-domain baseline in several settings.
\textit{Code is available at: \href{https://github.com/imigraz/SRCSM_Domain_Generalization}{https://github.com/imigraz/SRCSM\_Domain\_Generalization}}
\end{abstract}


\end{frontmatter}

\thispagestyle{firstpage}

\input{our_paper/introduction}

\input{our_paper/related_work}

\input{our_paper/method}

\input{our_paper/experiments}

\input{our_paper/discussion}

\input{our_paper/conclusion}

\section*{Funding Sources}

This research was funded in whole or in part by the Austrian Science Fund (FWF) 10.55776/PAT1748423. This work was further supported by grant I6540 from the Austrian Science Fund (FWF) and by grant FO999891133 from the Austrian Research Promotion Agency (FFG).

\bibliographystyle{elsarticle-num}
\bibliography{our_paper/references}

\end{document}

%% file: our_paper/defs.tex
\usepackage{euscript}
\usepackage{amssymb}
\usepackage{multirow}
\usepackage[misc]{ifsym}
\usepackage[super]{nth}
\usepackage{diagbox}
\usepackage{hhline}
\usepackage{subcaption}
\usepackage{xcolor}

\usepackage[
  colorlinks=true,
  linkcolor=black,   
  citecolor=black,   
  urlcolor=blue      
]{hyperref}

\usepackage[normalem]{ulem}

\usepackage{colortbl}

\newcommand{\lightvalue}{0.92}
\definecolor{lightgray}{rgb}{\lightvalue,\lightvalue,\lightvalue}

\usepackage{arydshln}

\usepackage{dcolumn}        
\newcolumntype{C}{D{.}{.}{2.2}} 

\usepackage{glossaries}
\glsdisablehyper

\newacronym{scn}{SCN}{Spatial-ConfigurationNet}
\newacronym{nnunet}{nnU-Net}{no-new-Net}
 
\newacronym{gdloss}{GD loss}{Generalized Dice Loss}
\newacronym{celoss}{CE loss}{Cross-Entropy Loss}

\newacronym[plural=CNNs,firstplural=Convolutional Neural Networks (CNNs)]{cnn}{CNN}{Convolutional Neural Network}
\newacronym[plural=ASMs,firstplural=Active Shape Models (ASM)]{asm}{ASM}{Active Shape Model}
\newacronym[plural=AAMs,firstplural=Active Appearance Models (AAM)]{aam}{AAM}{Active Appearance Model}
\newacronym{cca}{CCA}{connected component analysis}
\newacronym{mas}{MAS}{Multi-Atlas Segmentation}
\newacronym{mcd}{MCD}{Monte-Carlo Dropout}

\newacronym{iid}{{i.i.d.}}{{independent and identically distributed}}
\newacronym{tl}{TL}{Transfer Learning}
\newacronym{da}{DA}{Domain Adaptation}
\newacronym{dg}{DG}{Domain Generalization}
\newacronym{dr}{DR}{Domain Randomization}
\newacronym{sfda}{SFDA}{Source-Free Domain Adaptation}
\newacronym{tta}{TTA}{Test-time Adaptation}

\newacronym{saap}{SAAP}{Semantics-Aware Augmentation Pipeline}
\newacronym{lsa}{LSA}{Label-Specific Augmentation}
\newacronym{rna}{RNA}{RandConv Network Augmentation}
\newacronym{sm}{SM}{Source Matching}
\newacronym{cda}{CDA}{Conventional Data Augmentation}
\newacronym{src}{SRC}{Semantic-aware Random Convolution}

\newacronym{iamdg}{IAM}{Intensity Augmentation and Matching}
\newacronym{srcsm}{SRCSM}{Semantic-aware Random Convolution and Source Matching}

\newacronym{ct}{CT}{Computed Tomography}
\newacronym{mr}{MR}{Magnetic Resonance}
\newacronym{hu}{HU}{Hounsfield Units}

\newacronym{miccai}{MICCAI}{International Conference on Medical Image Computing and Computer-Assisted Intervention}
\newacronym{mmwhs}{MMWHS}{Multi-Modality Whole Heart Segmentation}
\newacronym{flare}{FLARE}{Fast and Low GPU Memory Abdominal Organ Segmentation}
\newacronym{amos}{AMOS}{Multi-Modality Abdominal Multi-Organ Segmentation}
\newacronym{bcv}{BCV}{Multi-Atlas Labeling Beyond the Cranial Vault}
\newacronym{bcvc}{BCV-C}{BCV-Cervix}
\newacronym{bcva}{BCV-A}{BCV-Abdomen}
\newacronym{chaos}{CHAOS}{Combined Healthy Abdominal Organ Segmentation}

\newacronym[plural=ROIs,firstplural=regions of interest (ROI)]{roi}{ROI}{region of interest}
\newacronym[plural=GPUs,firstplural=graphics processing units (GPU)]{gpu}{GPU}{graphics processing unit}

\newacronym{lv}{LV}{left ventricle}
\newacronym{rv}{RV}{right ventricle}
\newacronym{la}{LA}{left atrium}
\newacronym{ra}{RA}{right atrium}
\newacronym{myo}{MYO}{myocardium}
\newacronym{aa}{AA}{ascending aorta}
\newacronym{pa}{PA}{pulmonary artery}

\newacronym{cg}{CG}{central gland}
\newacronym{pz}{PZ}{peripheral zone}

\newacronym{li}{LI}{liver}
\newacronym{lk}{LK}{left kidney}
\newacronym{rk}{RK}{right kidney}
\newacronym{sp}{SP}{spleen}
\newacronym{st}{ST}{stomach}

\newacronym{bmc}{BMC}{Boston Medical Center}
\newacronym{runmc}{RUNMC}{Radboud University Nijmegen Medical Center}

\newacronym{dsc}{DSC}{Dice Similarity Coefficient}
\newacronym{asd}{ASSD}{Average Symmetric Surface Distance}
\newacronym{hd}{HD95}{Hausdorff Distance 95th Percentile}
\newacronym{ci}{CI}{confidence interval}
\newacronym{cd}{CD}{critical difference}

\newacronym{rsc}{RSC}{representation self-challenging}

\newacronym{ge}{GE}{General Electric}
\newacronym{siem}{Siem.}{Siemens}
\newacronym{phil}{Phil.}{Philips}

\newcommand{\cdfvar}{\mathcal{C}}
\newcommand{\srcvar}{\mathcal{S}}
\newcommand{\dstvar}{\mathcal{T}}

\newcommand{\cdfinvsrc}{\cdfvar_{\bar{\srcvar}}^{-1}}
\newcommand{\cdfdst}{\cdfvar_{\dstvar}}

\newcommand{\imagelettersmall}{x}

\newcommand{\image}{\mathbf{\imagelettersmall}}

\newcommand{\borderpixelmap}{\mathbf{m}}
\newcommand{\binarymap}{\bar{\mathbf{m}}}

\newcommand{\classlabelbig}{C}
\newcommand{\classlabelsmall}{c}

\newcommand{\borderpixelmapwlabel}{\borderpixelmap_{\classlabelsmall}}
\newcommand{\binarymapwlabel}{\binarymap_{\classlabelsmall}}

\newcommand{\rnanetletter}{g}
\newcommand{\rnanet}{\rnanetletter}

\newcommand{\rnaalpha}{\alpha^{\rnanetletter}}

\newlength\dunder
\newcommand{\doubleunderscore}{\rule{2\dunder}{0.4pt}} 

\newcommand{\papercsdg}{$^{a}$}
\newcommand{\paperirlsg}{$^{b}$}
\newcommand{\paperslaug}{$^{c}$}

\newcommand{\paperdcon}{$^{d}$}
\newcommand{\paperarfu}{$^{e}$}

\newcommand{\paperdesam}{$^{f}$}
\newcommand{\paperadami}{$^{g}$}
\newcommand{\papercthreer}{$^{h}$}
\newcommand{\papercuda}{$^{i}$}
\newcommand{\papermpscl}{$^{j}$}
\newcommand{\paperttas}{$^{k}$}
\newcommand{\paperhsd}{$^{l}$}


%% file: our_paper/introduction.tex
\section{Introduction}

Deep neural networks are known to underperform in the presence of domain shift, especially when test data have been generated from a vastly different distribution than the data used to train the network~\cite{Ben-David2006-ck,Torralba2011-gm}.
Domain shift has a strong impact in the context of medical imaging tasks, where the same anatomy can be captured with different imaging modalities, but models trained in one modality, e.g., \gls{mr}, may even fail completely when applied to another modality, e.g., \gls{ct}~\cite{AlBadawy2018-gi,Pooch2020-vz}.
Consequently, to achieve state-of-the-art performance, medical imaging applications require annotated data from all target domains, which is often prohibitively costly due to the expertise required to produce the annotations.

Single-source \gls{dg} methods~\cite{Zhou2023-jk,Wang2023-vz} seek to address this problem by training-time techniques that promote generalization of the network to data from unseen domains.
Many existing approaches rely on strong, randomized source data augmentation~\cite{zhang2020generalizing,Xu2020-vo,Choi2023-qy,Ouyang2023-xf,Su2023-du} or adversarial strategies~\cite{qiao2020learning,xu2022adversarial,yang2021adversarial,Chen2020-xe} to synthetically diversify the training data.
In doing so, these strategies aim to expand the model's coverage towards unseen test data distributions by making the model more robust towards synthetically introduced differences.
However, ensuring coverage of unseen test domains is inherently difficult.
As shown in our evaluation in Section~\ref{sec:quantitative_comparison_to_indomain}, existing methods are still far away from closing the performance gap between the source and target domains.

\input{figures/thale1}

To reduce this gap, we propose a single-source \gls{dg} method for semantic segmentation based on two key contributions:
First, we note that appearance differences between images acquired with different modalities are largely determined by the tissue type.
Consequently, augmentation strategies that are oblivious to tissue types are unlikely to produce appearance changes that are expected between source and target domain data.
We address this by introducing \glsreset{src}\gls{src}, where a distinct, highly non-linear random augmentation is applied to each region of source domain data that corresponds to a different ground truth label.
Second, we argue that it is much easier to shift test data towards the known distribution of the training data than to sufficiently augment the training data to ensure coverage of \textit{any} unknown target domain.
Driven by this observation, we propose to remap the intensity of test images at test-time in order to make them more similar to those of the training domain, which we call \glsreset{sm}\gls{sm}.
We combine these two contributions with conventional augmentation strategies and propose a single-source \gls{dg} method named \gls{srcsm}.
An overview of our method is provided in Fig.~\ref{fig1}.

To assess the performance of \gls{srcsm}, we perform an extensive evaluation including cross- modality and cross-center generalization on four abdominal, eight cardiac and six prostate domains in multiple settings, resulting in a large variety of source-target combinations.
Notably, we extend the common abdominal organ segmentation setting by using two additional, significantly larger domains: AMOS \gls{ct} and AMOS \gls{mr}~\cite{Ji2022-uv}.
Further, besides the widely used four label setting for whole-heart cardiac segmentation employed in literature, we also evaluate a more detailed setting that uses seven labels.
Lastly, to the best of our knowledge, we are the first to benchmark generalization in the challenging setting, where a model is trained on whole-heart \gls{ct} or whole-heart \gls{mr} data and evaluated on cardiac cine \gls{mr} data.
This setting results in a large and multifaceted domain gap, dramatically exceeding domain shifts evaluated by prior work.

In this massive set of experiments, \gls{srcsm} outperforms a wide range of existing methods in a vast majority of cases, setting a new state-of-the-art for cross-modality and cross-site single-source \gls{dg}.
Additionally, \gls{srcsm} even outperforms several \gls{da} and \gls{tta} approaches that have an advantage over single-source \gls{dg} methods as they adapt the model on target domain data.
Most notably, \gls{srcsm} significantly alleviates the impact of domain shift by achieving results that are close to or even in-line with the in-domain performance.
We summarize our contributions as follows:
\begin{itemize}
    \item We introduce \glsreset{src}\gls{src}, a highly nonlinear augmentation strategy that accounts for semantic labels.

    \item We propose to use an intensity quantile mapping called \glsreset{sm}\gls{sm} to align target domain images to the source domain at test-time.
    
    \item We extend the popular abdominal organ segmentation task for single-source \gls{dg} by adding a second \gls{ct} and a second \gls{mr} domain, leading to the most comprehensive evaluation on abdominal \gls{dg} reported so far.
    
    \item To the best of our knowledge, we are the first to evaluate generalization from whole-heart \gls{mr} and \gls{ct} data to diastolic and systolic cardiac cine \gls{mr} data.

    \item With our \gls{srcsm} approach, we establish a new state-of-the-art in abdominal organ, whole-heart and prostate segmentation, and achieve results that close in on or even match the in-domain baseline.
\end{itemize}

%% file: figures/thale1.tex
\begin{figure*}[t]
\includegraphics[width=\textwidth]{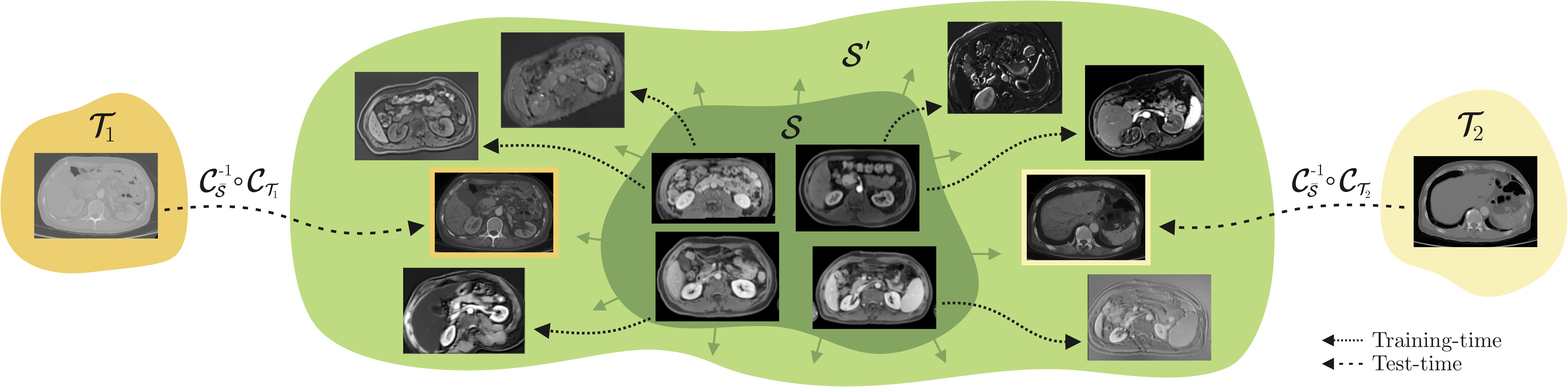}
\caption{
We propose SRCSM, a single-source cross-modality domain generalization approach that (1) during training, expands the source data distribution ($\srcvar$) through our Semantic-aware Random Convolution (SRC), and (2) at test-time, shifts images from unseen target domains ($\dstvar_1$ and $\dstvar_2$) towards the source domain ($\srcvar$) via our Source Matching (SM) strategy.
}
\label{fig1}
\end{figure*}

%% file: our_paper/related_work.tex
\section{Related Work}

In Section~\ref{sec:related_dg}, we review related techniques that adhere to the \gls{dg} scenario.
To put our method in a broader context, in Section~\ref{sec:related_da_tta} we highlight the differences between \gls{dg} and two other approaches of handling domain shift: \gls{da} and \gls{tta}.

\subsection{Domain Generalization}
\label{sec:related_dg}

In the \gls{dg} scenario~\cite{Zhou2023-jk,Wang2023-vz}, model parameters are exclusively updated based on source domain data and generalization is commonly achieved by diversifying the source domain.
\gls{dg} methods can be separated into two groups based on whether they employ one or require multiple source domains for training.
Multi-source domain generalization methods~\cite{Zhou2023-jk,Wang2023-vz} commonly aim to conflate several source data distributions such that the model may better align unseen target domain data to this conflated space.
In contrast, single-source \gls{dg} assumes that data from only one source domain is available~\cite{Zhou2023-jk,Wang2023-vz} which is inherently more challenging.
In this work, we focus on single-source \gls{dg} as it is more general and more relevant to medical imaging due to the scarcity of labeled data.
 
The palette of methods to promote generalization is broad and includes conventional~\cite{zhang2020generalizing,zhao2023augmentation} and adaptive data augmentation~\cite{huang2025ada}, dropping regions of the input image or features~\cite{DeVries2017-ad,Huang2020-lc}, randomly perturbing the computed features~\cite{yi2024hallucinated}, and mixing features computed for different object instances~\cite{Zhou2021-ie,chen2022maxstyle}.
While these approaches can serve as a good baseline in some settings, they are often insufficient in more challenging cross-modality scenarios.
To increase the efficacy of data augmentation, the authors of SLAug~\cite{Su2023-du} and PCSDG~\cite{jiang2025structure} proposed to remap the intensity of each semantically distinct region using a dedicated B\'{e}zier curve.
This, however, limits these approaches to intensity alterations that can be expressed by such curves.
In contrast, our method applies highly non-linear random convolutions to each semantic region, which allows a much larger variation of possible outcomes.
As shown in Section~\ref{sec:quantitative_comparison_related_work}, this lets our \gls{srcsm} systematically outperform SLAug~\cite{Su2023-du}, PCSDG~\cite{jiang2025structure} and other related augmentation methods in multiple settings.

Image augmentation was also used to promote feature alignment~\cite{yi2024hallucinated,peng2022semantic,wang2025hybrid}, or disentanglement of domain-specific and domain-invariant features~\cite{choi2021robustnet,niu2024irlsg,jiang2025structure} with dedicated loss functions.
The idea is that aligned or disentangled features display much less variation in response to the domain shift.
Our approach is much simpler:
Instead of aligning or disentangling features,
our method makes the model robust to domain shifts by diversifying the appearance of training images. 
The experiments in Section~\ref{sec:quantitative_comparison_related_work} show that \gls{srcsm} outperforms this line of approaches in a broad range of experiments.

Several techniques focus on adversarial domain augmentation~\cite{qiao2020learning,xu2022adversarial,yang2021adversarial,Chen2020-xe} to generate diverse synthetic training data.
However, it has been shown that the same effect can be achieved with augmentation based on random convolutions~\cite{Xu2020-vo,Ouyang2023-xf}, which is computationally much more efficient.

RandConv~\cite{Xu2020-vo} relies on a single randomly initialized convolution layer to augment the appearance of the source data.
Targeting object recognition in computer vision, the RandConv strategy was later extended to a shallow multi-layered network comprising a sequence of randomly initialized convolution layers~\cite{Choi2023-qy}. 
Concurrently, the same extension was also integrated in a similar way into domain generalization for medical image segmentation by~\cite{Ouyang2023-xf}, within their CSDG method.
Recently, the same approach was also investigated in~\cite{scholz2026random}, and adopted in combination with a contrastive learning strategy by DCON~\cite{wang2025hybrid}.
In contrast to approaches that augment the whole image with the same random convolution(s), \gls{srcsm} makes random convolutions semantic-aware by conditioning them on the semantic label of the augmented region.
ARFU~\cite{ren2025anatomically}, developed concurrently to our method, employs distinct random convolutions to each label similarly to us.
However, our evaluation in Section~\ref{sec:quantitative_comparison_related_work} confirms that our contributions result in considerable segmentation performance improvements compared to RandConv~\cite{Xu2020-vo}, CSDG~\cite{Ouyang2023-xf}, DCON~\cite{wang2025hybrid} and ARFU~\cite{ren2025anatomically}.

A completely separate line of single-source \gls{dg} approaches aims to fine-tune prompt-driven vision foundation models to turn them into automated task-specific models~\cite{zhang2023customized,gao2024desam,lin2024beyond}.
Our evaluation shows that the proposed \gls{srcsm} method outperforms these approaches by a large margin.

\subsection{Domain Adaptation and Test-time Adaptation}
\label{sec:related_da_tta}

In the Domain Adaptation (\gls{da}) scenario, data from the target domain is available during training.
\gls{da} approaches use this data to adapt the model for performance in the target domain~\cite{Guan2022-vr,Kumari2024-kl}.
This is in stark contrast to our \gls{dg} use case, where no target domain data is available at training-time and generalization is by design not limited to a previously known target domain.
One methodological similarity between our method and certain \gls{da} approaches~\cite{Ma2021-nw,Yaras2024-jr} is the use of histogram matching.
However, while the~\gls{da} methods modify the intensity of the source-domain training data to make the training images more similar to the target domain, \gls{srcsm} performs the matching in the opposite direction: The histogram of each target domain image is aligned to that of the source domain before generating predictions.

In Test-Time Adaptation (\gls{tta}), the model is fine-tuned on target domain data at test-time \cite{liang2025comprehensive,zhu2025improving}.
The disadvantage of this approach is that it is computationally expensive.
Moreover, updating the model on test data requires a heuristic stopping criterion to prevent overfitting to an unsupervised optimization target.
However, identifying generic stopping criteria that perform well on individual images even of the same target domain remains an open problem~\cite{zhu2025improving}.
By contrast, \gls{srcsm} does not adapt the model at test-time and induces virtually no computational overhead.

Even though \gls{da} and \gls{tta} approaches leverage target domain data to adapt the model while \gls{srcsm} does not, \gls{srcsm} outperforms several recent \gls{da} and \gls{tta} methods.
Notably, our key contributions, i.e., the semantic-aware random convolution and the source matching, could also be used to potentially boost performance of a \gls{da} or \gls{tta} approach.
However, in this paper, we adhere to the pure \gls{dg} scenario and leave combining \gls{srcsm} with \gls{da} or \gls{tta} techniques to future work.

%% file: our_paper/method.tex
\section{Method}
\label{sec:method}

\gls{srcsm} comprises two complementary components:
a data augmentation strategy aimed at extending the coverage of the network to unknown target domains,
and an intensity mapping routine to bring test images closer to the known source distribution.
We describe them in the following sections.

\subsection{Semantic-aware Random Convolution (SRC)}
\label{sec:training_time_strategies}

The point of departure for our method is the convolutional data augmentation~\cite{Xu2020-vo,Choi2023-qy,Ouyang2023-xf},
where input images are augmented with a shallow, convolutional network $g$.
Importantly, the network is never trained:
For every augmented image, its parameter vector $\theta$ is sampled from the standard normal distribution. 
This approach ensures a rich appearance variation, and 
convolutional data augmentation proved to be very effective.
However, it suffers from an important limitation:
It cannot reproduce the variation of relative contrast between different anatomical structures across different imaging modalities.
For example, when comparing the contrast between anatomic regions in the cardiac \gls{mr} image in Fig.~\ref{fig:lsa_motivation} (left), the left ventricle (red) appears similar to the right ventricle (green), while both are clearly distinguishable from the myocardium (cyan).
By contrast, in the \gls{ct} image (right), the right ventricle (green) is similar to the myocardium (cyan) and shows strong contrast to the left ventricle (red).
This type of domain shift cannot be simulated with common augmentation techniques, 
where two regions with the same input intensity are bound to receive the same intensity in the augmented image,
even if they correspond to different anatomical structures.

\input{figures/thale2}

To address this limitation, we propose \glsreset{src}\gls{src}.
It consists in applying distinct nonlinear augmentations to image regions corresponding to distinct semantic labels.
For each semantic label $\classlabelsmall\in\classlabelbig$, where $\classlabelbig$ is the set of anatomic labels, we generate a class-specific augmentation operator in the form of a convolutional neural network $g(\cdot,\theta_c)$, where the parameter vector $\theta_c$ is randomly drawn from the normal distribution.
Implementation details of $g(\cdot,\theta_c)$ are provided in Section~\ref{sec:implementation_details}.
In the naive version of \gls{src}, $g(\cdot,\theta_c)$ is used to augment the image region corresponding to the label $\classlabelsmall$ and represented by its binary ground truth mask $\binarymapwlabel$:
\begin{equation}
    \text{SRC}_\text{binary} (\image,\binarymap) = 
        \sum_{\classlabelsmall\in\classlabelbig}
	    \binarymapwlabel \odot \rnanet(\image, \theta_{\classlabelsmall}), \quad \theta_c \sim N(0,1),
\label{eq:sarc_simplified}
\end{equation}
where $\odot$ represents pixel-wise multiplication, and $N(0,1)$ is a standard normal distribution. 
However, we observed that images augmented with Eq.~\eqref{eq:sarc_simplified} often display amplified contrast around label borders.
This contrast hints at the location of the borders and leads to overfitting,
as shown by the ablation study that we present in Section~\ref{sec:ablation_study}.

\input{figures/thale3}

\input{figures/thale4}

We address this by blending the individual random convolution outputs close to the label borders.
To this end, we apply a 3D Gaussian kernel $G$ to each binary mask $\binarymapwlabel$,
yielding a smooth blending map 
\begin{equation}
    \borderpixelmapwlabel = \binarymapwlabel * G,
\end{equation}
where $*$ denotes the convolution operator.
In practice, every voxel $p$ in $\binarymapwlabel$ receives one label, such that \mbox{$\sum_{\classlabelsmall\in\classlabelbig} \binarymapwlabel[p] = 1$}.
Therefore, the smoothed map $\borderpixelmapwlabel$ also sums to one over the label dimension.
Thus, $\borderpixelmapwlabel$ can be used for blending the individual random convolution outputs, yielding the proposed \gls{src} strategy:
\begin{equation}
    \text{SRC} (\image,\borderpixelmap) = 
        \sum_{\classlabelsmall\in\classlabelbig}
	    \borderpixelmapwlabel \odot \rnanet(\image, \theta_{\classlabelsmall}).
\label{eq:src}
\end{equation}
This strategy ensures smooth transitions between different image regions, crucial for generalizing to unseen target domains.

To ensure a full range of augmentation capacities, we complement \gls{src} with geometric augmentation: 
We employ translation, rotation, scaling and elastic deformation.
The complete augmentation approach is shown in Figure~\ref{fig:intensity_augmentation_schematic} and exemplary images before and after applying \gls{src} are shown in Figure~\ref{fig:images_src}.

\input{figures/thale5}

\subsection{Source Matching (SM)}
\label{sec:test_time_strategies}

Even though our augmentation approach is already effective in reducing the performance gap between source and target domain data, some gap in performance remains, as shown in our ablation study in Section~\ref{sec:ablation_study}.
To further reduce the remaining domain gap, we complement the proposed source data augmentation techniques by shifting images from the target domain towards the source domain at test-time.

To that end, we propose to use intensity quantile mapping which we call \glsreset{sm}\gls{sm}.
\gls{sm} transforms the individual histogram of each target domain image such that it aligns with the average histogram of the source domain dataset.
Note that \gls{sm} is designed to shift target images such that they coarsely approximate the source data distribution, an approximation which need not be perfect due to the complementary influence of \gls{src}.
%
%
Formally, each voxel with intensity $v$ of a given test image is assigned a new value:
\begin{equation}
    \text{SM} (v) = \cdfinvsrc \circ \cdfdst (v) ,
    \label{eq:sm}
\end{equation}
where $\cdfdst$ denotes the cumulative probability density function of voxel intensities in the given test image, and $\cdfinvsrc$ is the average quantile function of voxel intensities in the source domain.
In other words, $\cdfdst$ maps the intensity $v$ to the probability that a randomly selected voxel of the \emph{test} image has intensity lower or equal to $v$.
Then, $\cdfinvsrc$ maps that probability for $v$ to the corresponding voxel intensity of the \emph{training} domain.

In practice, we implement both mappings using cumulative, normalized histograms.
We approximate $\cdfdst(v)$ with the cumulative histogram of voxel intensities in the given test image.
Then, given a probability $p=\cdfdst(v)$, we compute $\cdfinvsrc(p)$ using the average cumulative histogram of all images in the source dataset.
Specifically, we denote the center of the $i$-th bin of the histogram by $c_{\bar{\srcvar}}[i]$ and the value of that bin by $h_{\bar{\srcvar}}[i]$, and define 
\begin{equation}
\cdfinvsrc (p)  = c_{\bar{\srcvar}}[{i^*}] \quad\text{where}\quad i^* = \arg \min_{i} (h_{\bar{\srcvar}}[i] - p)^2.
\end{equation}
The average histogram of the source domain $h_{\bar{\srcvar}}$ is only computed once during training and stored alongside the model for later use.
Exemplary images before and after applying \gls{sm} are shown in Figure~\ref{fig:images_sm}.
Our ablation study in Section~\ref{sec:ablation_study} shows that \gls{sm} achieves performance improvements that are complementary to \gls{src}.
Importantly, \gls{sm} is very light-weight and efficient, imposing almost no computational overhead.

%% file: figures/thale2.tex
\begin{figure}[t]
\includegraphics[width=\columnwidth]{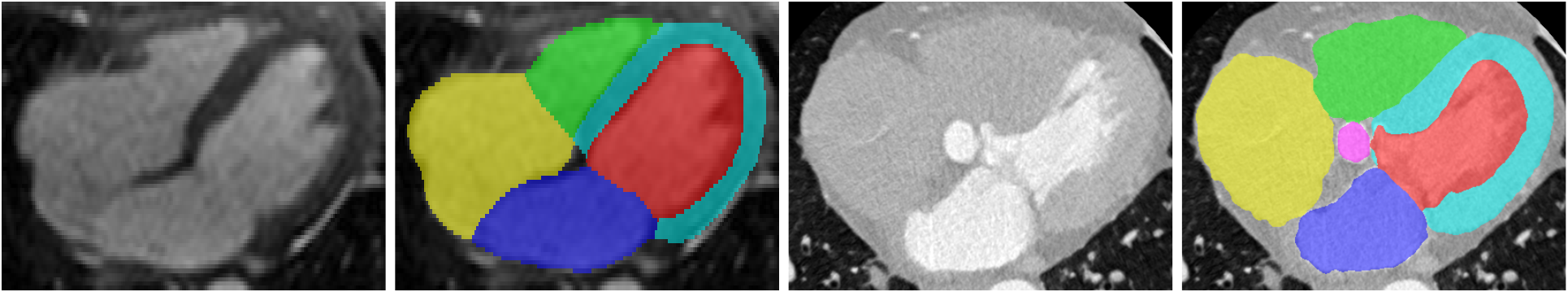}
\caption{
Contrast between anatomical structures strongly depends on the imaging modality.
Left: a cardiac MR and its ground truth annotation;
Right: the same anatomy visible in a cardiac CT.
Note the differences between MR and CT in relative intensity of the left ventricle (red), the right ventricle (green), and the myocardium (cyan).
}
\label{fig:lsa_motivation}
\end{figure}

%% file: figures/thale3.tex
\begin{figure*}[t]
\includegraphics[width=\textwidth]{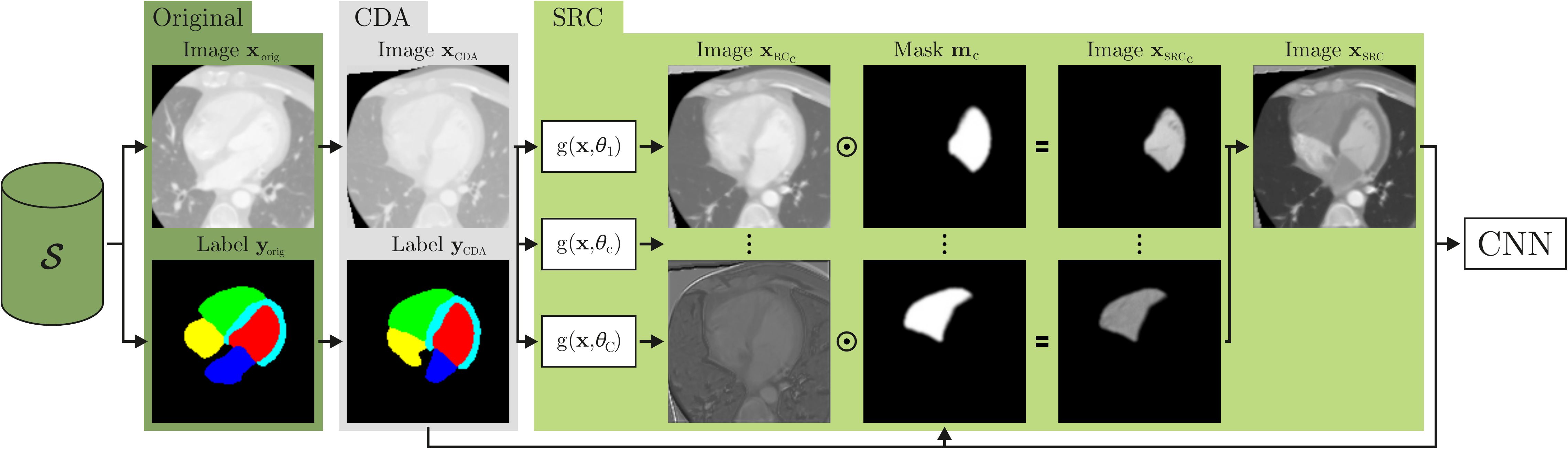}
\caption{
Our domain generalization approach:
Images from source domain $\srcvar$ are first augmented using conventional data augmentation (CDA).
Then, our Semantic-aware Random Convolution (SRC) strategy applies a distinct nonlinear intensity augmentation for each available semantic label $\classlabelsmall\in\classlabelbig$.
After that, each augmented image is masked using its corresponding smoothed map $\borderpixelmapwlabel$, before being recombined yielding the final SRC-augmented image.
}
\label{fig:intensity_augmentation_schematic}
\end{figure*}

%% file: figures/thale4.tex
\begin{figure*}[t]
\includegraphics[width=\textwidth]{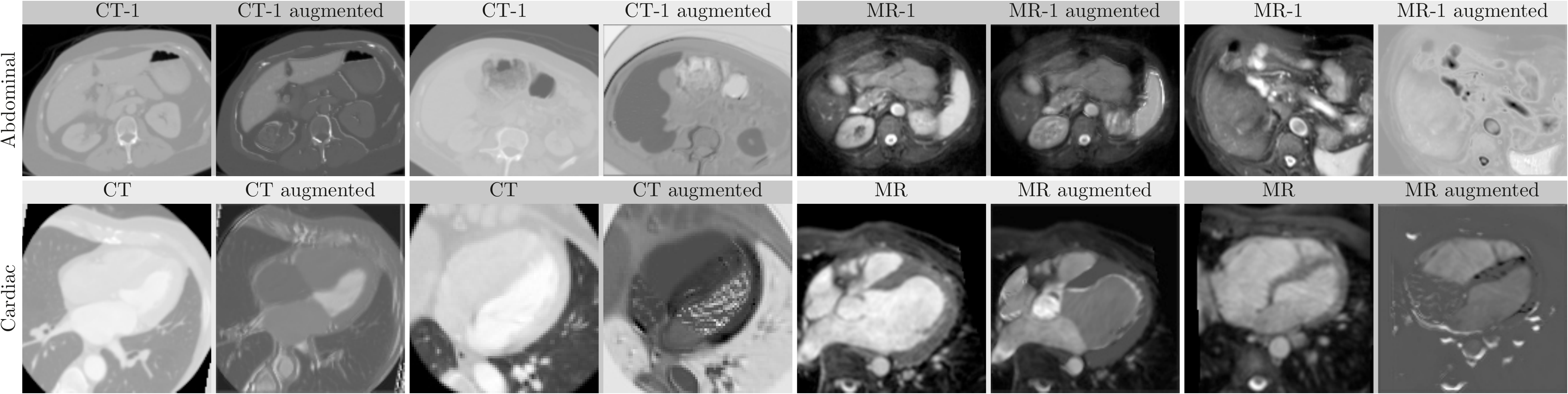}
\caption{
Exemplary abdominal (row 1) and cardiac (row 2) images are shown before (cols: 1, 3, 5, 7) and after (cols: 2, 4, 6, 8) applying the proposed Semantic-aware Random Convolution (SRC).
Image contrast was adjusted for better visualization.
}
\label{fig:images_src}
\end{figure*}

%% file: figures/thale5.tex
\begin{figure*}[t]
\includegraphics[width=\textwidth]{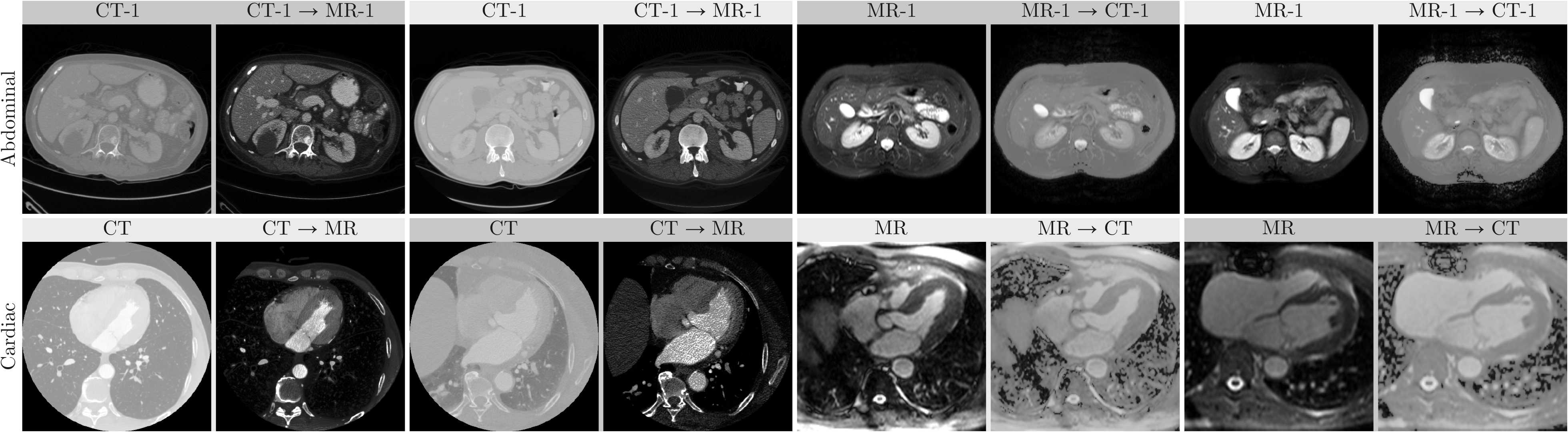}
\caption{
Exemplary abdominal (row 1) and cardiac (row 2) images are shown before (cols: 1, 3, 5, 7) and after (cols: 2, 4, 6, 8) applying the proposed Source Matching (SM).
Image contrast was adjusted for better visualization.
}
\label{fig:images_sm}
\end{figure*}

%% file: our_paper/experiments.tex
\section{Experimental Setup}
\label{section3}

\subsection{Datasets}
\label{sec:datasets}

Our comprehensive evaluation relies on a large variety of different datasets and setups.
A description is provided in the following, a tabular overview is presented in Table~\ref{tab:dataset_table}.

\subsubsection*{Cross-modality Abdominal Organ Segmentation}

We employ four domains for abdominal organ segmentation:
\begin{itemize}
    \item \textbf{BCV (CT-1)}~\cite{Landman2015-hu} consists of 30 labeled \gls{ct} images. We use the same train-val-test split, i.e., 70-10-20, as CSDG~\cite{Ouyang2023-xf} in Tables~\ref{tab:related_work_abdominal_csdg_setup}, \ref{tab:abdominal_detailed} and \ref{tab:ablation}.
    For Table~\ref{tab:related_work_abdominal_desam_setup}, we use the same 90-10 train-test split as DeSAM~\cite{gao2024desam}.
    Both setups employ all images when used as out-of-domain test set.
    For in-domain comparison, we use the 20\% test split from CSDG.
    
    \item \textbf{CHAOS (MR-1)}~\cite{Kavur2021-xj} includes 20 labeled \gls{mr} images. 
    The remaining information is identical to CT-1.
    
    \item \textbf{AMOS (CT-2)}~\cite{Ji2022-uv} contains 300 labeled \gls{ct} scans which are already separated into an official training and validation set.
    While an official test set also exists, it is not publicly available with ground truth annotations.
    For consistency with the other datasets used in this work, we refer to the samples constituting the official validation set as the test set in this work.
    The same applies for AMOS (MR-2).
    After excluding cases with any of the four considered labels missing, 197 scans of the training set and 98 scans of the test set remain.
    When used as out-of-domain test set, we employ all 295 remaining scans.
    For in-domain comparison, we use the 98 scans from the test set.
            
    \item \textbf{AMOS (MR-2)}~\cite{Ji2022-uv} also includes 60 labeled \gls{mr} scans which constitute a training set of 39 scans and a test set of 20 scans after excluding cases with any of the considered labels missing.
    As out-of-domain test set, all 59 cases are used.
    In-domain comparisons are performed using the test set of 20 scans.

\end{itemize}
We follow previous work and evaluate segmentation of four organs:
liver, left kidney, right kidney and spleen.

\subsubsection*{Cross-modality Whole-Heart Segmentation}

We use the following domains in heart segmentation experiments:
\begin{itemize}
    \item \textbf{MMWHS (CT)}~\cite{Zhuang2019-uj} includes 20 labeled \gls{ct} scans.
    For our evaluation in Table~\ref{tab:related_work_mmwhs}, we use the same train-test split, i.e., 80-20, as related work which was popularized in~\cite{Chen2020-kh}.
    Importantly, this setup only considers four labels: \gls{lv}, \gls{la}, \gls{myo} and \gls{aa}, and only employs four images as out-of-domain test set.
    In-domain comparisons in the 4-label setup have been evaluated on the same four images that constitute the out-of-domain test set and therefore trained on the 16 remaining scans of the same domain.
    For a more detailed evaluation of the MMWHS dataset in Table~\ref{tab:related_work_mmwhs_7labels}, we use all seven labels and employ all 20 images for training and all 20 images from the other domain as out-of-domain test set.
    In-domain comparisons in the 7-label setup are performed on the 40 test scans with encrypted labels, which requires using the evaluation script provided by the challenge organizers.
        
    \item \textbf{MMWHS (MR)}~\cite{Zhuang2019-uj} consists of 20 labeled \gls{mr} scans.
    The remaining information is identical to MMWHS CT.

\end{itemize}
Overall, seven labels are available: \gls{lv}, \gls{rv}, \gls{la}, \gls{ra}, \gls{myo}, \gls{aa} and \gls{pa}.

\subsubsection*{Cardiac Cine MR as Target Domain}

We also consider bridging the domain gap from whole-heart to cardiac cine \gls{mr} data by employing \textbf{M\&Ms-2}~\cite{martin2023deep}.
This dataset includes short-axis scans of 360 subjects with ground truth segmentations of the diastolic and systolic phase of the cardiac cycle for which three labels are available: \gls{lv}, \gls{rv} and \gls{myo}.
We only considered the 160 subjects that comprise the official test set, however, excluded three cases for which the orientation information was erroneous.
The dataset includes scans obtained with different scanner hardware: 
General Electric ({\bf GE}), Philips ({\bf Phil.}) and Siemens ({\bf Siem.}).
With three scanner manufacturers and two phases of the cardiac cycle (\emph{Systolic} and \emph{Diastolic}), this dataset contributes six additional domains.

\input{tables/thale.t1}

\subsubsection*{Cross-center Prostate Segmentation}

For prostate segmentation, we employ six popular domains from three datasets: NCI-ISBI13~\cite{bloch2015nci}, I2CVB~\cite{lemaitre2015computer} and PROMISE12~\cite{litjens2014evaluation} in the version provided by~\cite{chen2022maxstyle,gao2024desam}.
To compare to related work, we use all available data from the respective source domain when comparing to HSD~\cite{yi2024hallucinated} in Table~\ref{tab:iam_related_work_prostate_hsd_setup}, and the same 90-10 train-test split as DeSAM~\cite{gao2024desam} in Table~\ref{tab:iam_related_work_prostate_desam_setup}. 
Out-of-domain test sets include all available data of the respective domain for both setups.
In-line with related work, we refer to these domains as A, B, C, D, E and F in the same order, which is:
RUNMC/ISBI (A),
BMC/ISBI1.5 (B),
I2CVB (C),
UCL (D),
BIDMC (E) and
HK (F).
The domains include 30, 30, 19, 13, 12 and 12 \gls{mr} scans, respectively.
Further imaging related details on the individual domains can be found in \cite{litjens2014evaluation,liu2020ms}.
For in-domain comparisons, we respectively use the 10\% validation splits from DeSAM.

\subsection{Implementation Details}
\label{sec:implementation_details}

\subsubsection*{Data Processing}
All our experiments are performed in 3D, images are resampled to have an isotropic spacing and we use a predefined landmark such that the approximated center of the region of interest for segmentation is in the center of the image.
Following the same strategy as in~\cite{Thaler2025}, we selected the resampling parameters as such that they tightly fit to the region of interest of all images in order to maximize the physical resolution.
Therefore, we resample abdominal data to $160 \times 128 \times 160$ voxel with $2.4 \text{ mm}$ spacing, for cardiac data we use $128 \times 128 \times 128$ voxel with $1.5 \text{ mm}$ spacing, and for prostate segmentation we employ $128 \times 128 \times 128$ voxel with $0.75 \text{ mm}$ spacing.

\subsubsection*{Network Architecture and Training}
The employed network architecture is similar to U-Net~\cite{Ronneberger2015-ih} with a contracting and expanding path consisting of five levels.
At each level we use two convolutions with $64$ filters and an intermediate dropout layer~\cite{Srivastava2014-rv} with a dropout rate of $0.1$.
After convolution layers we employ leaky ReLU as activation function with alpha $0.1$.
We use average pooling as well as linear upsampling layers, and padding is set to 'same' to preserve the spatial size of the convolution input.
Each model is trained for $50,000$ iterations with a batch size of $1$ and we use generalized Dice as a loss function.
As optimizer we use Adam~\cite{Kingma2015-fb} with a learning rate of $5e^{-5}$.
Convolution kernels are initialized using the He initializer~\cite{He2015-hz} with an L2 kernel regularizer.
We employ temporal ensembling by computing the exponential moving average~\cite{Laine2017-zh} for each weight with a decay of $0.999$.

\input{tables/thale.t2}

\input{tables/thale.t3}

\subsubsection*{Conventional Data Augmentation (CDA)}
Same as in~\cite{Thaler2025augmentation}, we augment training data in 3D using translation ($\pm 20$ voxel), rotation ($\pm 0.35$ radians), scaling ($[0.8, 1.2]$) and elastic deformation (eight grid nodes per dimension, deformation values are sampled from $\pm 15$ voxels).
\gls{ct} images are expected in \gls{hu}, normalized by dividing intensity values by 2048, and are clipped to the range $[-1, 1]$ to protect the model against extremely low or high intensity values introduced, e.g., by device artifacts.
For robust \gls{mr} normalization, the \nth{10} and \nth{90} percentile are linearly normalized to the range $[-1, 1]$, respectively.
After normalization, a random intensity shift ($\pm 0.2$) as well as an intensity scaling with a factor between $[0.8, 1.2]$ (\gls{ct}) or $[0.6, 1.4]$ (\gls{mr}) are applied globally.

\subsubsection*{Semantic-aware Random Convolution (SRC)}

The convolutional neural network $g(\cdot,\theta_c)$ used for data augmentation consists of a series of four convolutional layers which are randomly re-initialized whenever a new image is augmented.
The feature maps of intermediate layers consist of two channels, while the output of the last layer consists of one channel such that the augmented image is consistent to the input.
Each convolution layer is followed by a leaky ReLU activation with the alpha parameter set to 0.1.
The kernel size of each random convolution layer is sampled to be either one or three whenever it is re-initialized.
The 3D Gaussian kernel $G$ uses $\sigma = 1$ in voxel and a kernel size of $5 \times 5 \times 5$ voxels.
Image blending is performed using a uniformly drawn $\rnaalpha \sim \mathcal{U}(0, 1)$.
We follow previous work~\cite{Hendrycks2019-ls,Xu2020-vo,Ouyang2023-xf} and linearly interpolate the output of the random convolution with the original image, to partially reduce the effect of the augmentation.
To avoid exploding gradients, we also re-normalize the augmented image to retain the same Frobenius norm as the original image.

\subsubsection*{Source Matching (SM)}
Before applying \gls{sm}, we clip \gls{ct} data, which is expected to be defined in Hounsfield units, to the range [-1023, 1024].
Since intensity ranges for \gls{mr} data are not well-defined, we robustly normalize them as such that their minimum value remains 0, while the $0.9$ percentile is scaled to 2047 before clipping them to [0, 2047].
This clipping does not remove important information:
Voxels clipped to the minimum are predominantly located outside the body, while voxels clipped to the maximum are typically outliers, e.g., device artifacts in \gls{ct} or surface coil artifacts in \gls{mr}.

\subsection{Baselines}

Our comprehensive evaluation on the various datasets involves a large variety of baseline approaches from related literature on \gls{dg}, \gls{da} and \gls{tta}.
\gls{dg} approaches include
RSC~\cite{Huang2020-lc},
MixStyle \cite{Zhou2021-ie},
MaxStyle~\cite{chen2022maxstyle},
AdvBias~\cite{Chen2020-xe},
Cutout~\cite{DeVries2017-ad},
RandConv~\cite{Xu2020-vo},
ISW~\cite{choi2021robustnet},
ATSRL~\cite{yang2021adversarial},
BigAug~\cite{zhang2020generalizing},
PCSDG~\cite{jiang2025structure},
M-ADA~\cite{qiao2020learning},
ADS~\cite{xu2022adversarial},
AugSeg~\cite{zhao2023augmentation},
IRLSG~\cite{niu2024irlsg},
HSD~\cite{yi2024hallucinated} in their CNN and transformer variant,
ARFU~\cite{ren2025anatomically},
DCON~\cite{wang2025hybrid},
CSDG~\cite{Ouyang2023-xf} and
SLAug~\cite{Su2023-du}.
\gls{dg} approaches based on prompt-driven vision-foundation models include
MedSAM~\cite{ma2024segment},
SAMed~\cite{zhang2023customized},
SAMUS~\cite{lin2024beyond} as employed in \cite{wang2025hybrid}, and
DeSAM~\cite{gao2024desam} in their whole box (DeSAM-B) and grid points (DeSAM-P) mode.
As \gls{da} approaches, we employ
CycleGAN~\cite{Zhu2017-sj},
SIFA~\cite{Chen2020-kh},
SynthMix~\cite{zhang2023synthmix},
CUDA~\cite{Du2022-xe},
MPSCL~\cite{Liu2022-zd} and
C3R~\cite{ding2025c3r}.
Finally, for \gls{tta}, we use
Prior AdaEnt~\cite{Bateson2020-gl},
AdaMI~\cite{Bateson2022-bv},
Tent~\cite{Wang2021-ho},
TTAS$_\EuScript{RC}$~\cite{Bateson2022-gh},
TTT~\cite{sun2020test},
TTST~\cite{Karani2021-hh} and
TASD~\cite{liu2022single}.

Related work scores for SLAug and CSDG indicated by $^{\dagger}$ were produced using the original authors code.
We employed the same data preprocessing for \gls{ct} and \gls{mr} data as used in their code, respectively.
For whole-heart experiments, we increased their upper thresholding parameter used for \gls{ct} data from $275$ to $675$ Hounsfield units to ensure that no foreground region of whole-heart \gls{ct} scans is removed.
Due to the large size and therefore excessive training time of \gls{ct}-2, we reduced the number of epochs from 2,000 to 1,000 when training on this dataset.
The In-Domain baselines were trained on images of the target domain and only employ conventional data augmentation.

\subsection{Evaluation and Statistical Comparison}
\label{sec:evaluation_and_statistical_comparison}

All quantitatively evaluated experiments using our method as well as the reproduced experiments using SLAug~\cite{Su2023-du}, CSDG~\cite{Ouyang2023-xf} and In-Domain baselines were repeated three times to address the stochastic nature of model predictions.
Standard deviations were computed over all images of the respective test set and their repetitions, unless otherwise stated in the table caption; for reported mean scores, we computed standard deviations after computing the mean over labels.
The test set size employed in each experiment is stated in the respective table caption to allow the computation of \glspl{ci}.
All \glspl{ci} are computed with a confidence level of 95\% and under the normality assumption.
Further, we performed a statistical analysis of multiple dataset generalization by considering three algorithms, i.e., the proposed \gls{srcsm}, as well as SLAug and CSDG.
To test the null hypothesis of these algorithms performing equally, we employed a total of 14 different domain generalization experiments, comprising all combinations of abdominal and cardiac 7-label datasets reported in Tables~\ref{tab:abdominal_detailed} and~\ref{tab:related_work_mmwhs_7labels}.
For this test, we used the Autorank package~\cite{Herbold2020-ea} as a Python wrapper for statistical comparisons over multiple datasets~\cite{Demsar2006-zu}.
Given a family-wise significance level with $p = 0.05$, Autorank analyzes normality of the populations (algorithms) and chooses appropriate tests for the existence of differences in the central tendency.
If any difference exists, it also chooses a post-hoc test for pairwise comparisons, resulting in a \gls{cd} that defines if differences between algorithms are statistically significant.

\subsection{Visualization of Intermediate Features}

In addition, we exemplarily assess the similarity of the intermediate feature representation of source and target domain data in Fig.~\ref{fig:tsne_all} to allow visually inspecting the remaining domain gap.
To this end, intermediate feature representations are obtained from the penultimate layer of the respective models in our ablation study based on the abdominal \gls{mr}-1 $\rightarrow$ \gls{ct}-1 experiment.
Due to the high dimensionality of the intermediate feature representation, we employ t-SNE~\cite{van2008visualizing} to reduce feature vectors from 64 to two dimensions.
One property of t-SNE is the preservation of local neighborhoods, i.e., points that are close to one another in the high-dimensional feature space are also close to one another in the low-dimensional feature space.
To reduce the computational complexity of t-SNE, we sample 6,000 points per label uniformly from all images in the source \textit{and} target domain, respectively.
The corresponding label for each point is identified directly from the ground truth segmentation.
Importantly, for each model, we apply t-SNE jointly to all sampled points from the source \textit{and} target domain such that their two-dimensional representations align to one another, allowing a side-by-side comparison.
For better visualization and comparison purposes in Fig.~\ref{fig:tsne_all}, we provide separate plots for source points (in-domain, col~1), target points (out-of-domain, col~2) and both superimposed onto one another (col~3).
The colors in col~1-2 represent the different foreground labels, namely, liver (red), left kidney (green), right kidney (blue) and spleen (yellow) and are plotted with transparency to convey point density.
The colors in col~3 represent source (magenta) and target points (cyan).
Moreover, in col~4-5, we provide exemplary out-of-domain images to complement the t-SNE plots, where contour lines represent the ground truth and colored semi-transparent areas represent predictions.

%% file: tables/thale.t1.tex
\begin{table*}[t]
\centering
\footnotesize
\resizebox{\textwidth}{!}{%
\begin{tabular}{| c | l | c|c|c|c|c|c|c|}
\hline

\multicolumn{2}{|c|}{\multirow{2}{*}{\textbf{Dataset}}} & \multicolumn{1}{c|}{\multirow{2}{*}{\textbf{Setup}}} & \multicolumn{1}{c|}{\multirow{2}{*}{\textbf{Total}}} & \multicolumn{1}{c|}{\multirow{2}{*}{\textbf{Train}}} & \multicolumn{1}{c|}{\multirow{2}{*}{\textbf{Val}}} & \multicolumn{2}{c|}{\multirow{1}{*}{\textbf{Test}}} & \multicolumn{1}{c|}{\multirow{2}{*}{\textbf{Used in}}} \\

\cline{7-8}

\multicolumn{2}{|c|}{\multirow{1}{*}{}} & & & & & \textbf{In-Domain} & \textbf{Out-of-Domain} & \\

\hline

\multirow{6}{*}{\textit{\rotatebox{90}{Abdominal}}}
& \multicolumn{1}{l|}{\multirow{2}{*}{\textbf{BCV (CT-1)}~\cite{Landman2015-hu}}} & as in CSDG~\cite{Ouyang2023-xf} & 30 & 21 (70\%) & 3 (10\%) & 6 (20\%) & 30 (100\%) & Tables~\ref{tab:related_work_abdominal_csdg_setup}, \ref{tab:abdominal_detailed}, \ref{tab:in_domain_comparison} and \ref{tab:ablation} \\
& & as in DeSAM~\cite{gao2024desam} & 30 & 27 (90\%) & -- & 3 (10\%) & 30 (100\%) & Table~\ref{tab:related_work_abdominal_desam_setup} \\

\cline{2-9}

& \multicolumn{1}{l|}{\multirow{2}{*}{\textbf{CHAOS (MR-1)}~\cite{Kavur2021-xj}}} & as in CSDG~\cite{Ouyang2023-xf} & 20 & 14 (70\%) & 2 (10\%) & 4 (20\%) & 20 (100\%) & Tables~\ref{tab:related_work_abdominal_csdg_setup}, \ref{tab:abdominal_detailed}, \ref{tab:in_domain_comparison} and \ref{tab:ablation} \\
& & as in DeSAM~\cite{gao2024desam} & 20 & 18 (90\%) & -- & 2 (10\%) & 20 (100\%) & Table~\ref{tab:related_work_abdominal_desam_setup} \\

\cline{2-9}

& \textbf{AMOS (CT-2)}~\cite{Ji2022-uv} & as in~\cite{Ji2022-uv} & 295$^{\dagger}$ & 197$^{\dagger}$ (67\%) & -- & 98$^{\dagger}$ (33\%) & 295$^{\dagger}$ (100\%) & Tables~\ref{tab:related_work_abdominal_csdg_setup}, \ref{tab:abdominal_detailed} and \ref{tab:in_domain_comparison} \\

\cline{2-9}

& \textbf{AMOS (MR-2)}~\cite{Ji2022-uv} & as in~\cite{Ji2022-uv} & 59$^{\dagger}$ & 39$^{\dagger}$ (67\%) & -- & 20$^{\dagger}$ (33\%) & 59$^{\dagger}$ (100\%) & Tables~\ref{tab:related_work_abdominal_csdg_setup}, \ref{tab:abdominal_detailed} and \ref{tab:in_domain_comparison} \\

\hline
\hline

\multirow{8}{*}{\textit{\rotatebox{90}{Cardiac}}}

& \textbf{MMWHS (CT)}~\cite{Zhuang2019-uj} (4 labels) & as in SIFA~\cite{Chen2020-kh} & 20 & 16 (80\%) & -- & 4 (20\%) & 4 (20\%) & Table~\ref{tab:related_work_mmwhs} \\
& \textbf{MMWHS (MR)}~\cite{Zhuang2019-uj} (4 labels) & as in SIFA~\cite{Chen2020-kh} & 20 & 16 (80\%) & -- & 4 (20\%) & 4 (20\%) & Table~\ref{tab:related_work_mmwhs} \\

\cline{2-9}

& \textbf{MMWHS (CT)}~\cite{Zhuang2019-uj} & as in~\cite{Zhuang2019-uj} & 20 & 20 (100\%) & -- & 40$^{\ddagger}$ & 20 (100\%) & Tables~\ref{tab:related_work_mmwhs_7labels}, \ref{tab:in_domain_comparison}, \ref{tab:ablation} \\
& \textbf{MMWHS (MR)}~\cite{Zhuang2019-uj} & as in~\cite{Zhuang2019-uj} & 20 & 20 (100\%) & -- & 40$^{\ddagger}$ & 20 (100\%) & Tables~\ref{tab:related_work_mmwhs_7labels}, \ref{tab:in_domain_comparison}, \ref{tab:ablation} \\

\cline{2-9}

& \textbf{M\&Ms-2}~\cite{martin2023deep} All & as in~\cite{martin2023deep} & 351$^{\mathsection}$ & 156$^{\mathsection}$ (44\%) & 38$^{\mathsection}$ (11\%) & 157$^{\mathsection}$ (45\%) & 157$^{\mathsection}$ (45\%) & Table~\ref{tab:whs_to_cine} \\
\cline{2-9}
& \textbf{M\&Ms-2}~\cite{martin2023deep} GE & as in~\cite{martin2023deep} & 53$^{\mathsection}$ & 26$^{\mathsection}$ (49\%) & 8$^{\mathsection}$ (15\%) & 19$^{\mathsection}$ (36\%) & 19$^{\mathsection}$ (36\%) & Table~\ref{tab:whs_to_cine} \\
& \textbf{M\&Ms-2}~\cite{martin2023deep} Phil. & as in~\cite{martin2023deep} & 88$^{\mathsection}$ & 64$^{\mathsection}$ (73\%) & 14$^{\mathsection}$ (16\%) & 10$^{\mathsection}$ (11\%) & 10$^{\mathsection}$ (11\%) & Table~\ref{tab:whs_to_cine} \\
& \textbf{M\&Ms-2}~\cite{martin2023deep} Siem. & as in~\cite{martin2023deep} & 210$^{\mathsection}$ & 66$^{\mathsection}$ (31\%) & 16$^{\mathsection}$ (8\%) & 128$^{\mathsection}$ (61\%) & 128$^{\mathsection}$ (61\%) & Table~\ref{tab:whs_to_cine} \\

\hline
\hline

\multirow{12}{*}{\textit{\rotatebox{90}{Prostate}}}

& \textbf{RUNMC/ISBI (A)}~\cite{bloch2015nci} & as in HSD~\cite{yi2024hallucinated} & 30 & 30 (100\%) & -- & -- & -- & Table~\ref{tab:iam_related_work_prostate_hsd_setup} \\
& \textbf{BMC/ISBI1.5 (B)}~\cite{bloch2015nci} & as in HSD~\cite{yi2024hallucinated} & 30 & -- & -- & -- & 30 (100\%) & Table~\ref{tab:iam_related_work_prostate_hsd_setup} \\
& \textbf{I2CVB (C)}~\cite{lemaitre2015computer} & as in HSD~\cite{yi2024hallucinated} & 19 & -- & -- & -- & 19 (100\%) & Table~\ref{tab:iam_related_work_prostate_hsd_setup} \\
& \textbf{UCL (D)}~\cite{litjens2014evaluation} & as in HSD~\cite{yi2024hallucinated} & 13 & -- & -- & -- & 13 (100\%) & Table~\ref{tab:iam_related_work_prostate_hsd_setup} \\
& \textbf{BIDMC (E)}~\cite{litjens2014evaluation} & as in HSD~\cite{yi2024hallucinated} & 12 & -- & -- & -- & 12 (100\%) & Table~\ref{tab:iam_related_work_prostate_hsd_setup} \\
& \textbf{HK (F)}~\cite{litjens2014evaluation} & as in HSD~\cite{yi2024hallucinated} & 12 & -- & -- & -- & 12 (100\%) & Table~\ref{tab:iam_related_work_prostate_hsd_setup} \\

\cline{2-9}

& \textbf{RUNMC/ISBI (A)}~\cite{bloch2015nci} & as in DeSAM~\cite{gao2024desam} & 30 & 27 (90\%) & -- & 3 (10\%) & 30 (100\%) & Tables~\ref{tab:iam_related_work_prostate_desam_setup} and \ref{tab:in_domain_comparison} \\
& \textbf{BMC/ISBI1.5 (B)}~\cite{bloch2015nci} & as in DeSAM~\cite{gao2024desam} & 30 & 27 (90\%) & -- & 3 (10\%) & 30 (100\%) & Tables~\ref{tab:iam_related_work_prostate_desam_setup} and \ref{tab:in_domain_comparison} \\
& \textbf{I2CVB (C)}~\cite{lemaitre2015computer} & as in DeSAM~\cite{gao2024desam} & 19 & 17 (90\%) & -- & 2 (10\%) & 19 (100\%) & Tables~\ref{tab:iam_related_work_prostate_desam_setup} and \ref{tab:in_domain_comparison} \\
& \textbf{UCL (D)}~\cite{litjens2014evaluation} & as in DeSAM~\cite{gao2024desam} & 13 & 11 (90\%) & -- & 2 (10\%) & 13 (100\%) & Tables~\ref{tab:iam_related_work_prostate_desam_setup} and \ref{tab:in_domain_comparison} \\
& \textbf{BIDMC (E)}~\cite{litjens2014evaluation} & as in DeSAM~\cite{gao2024desam} & 12 & 10 (90\%) & -- & 2 (10\%) & 12 (100\%) & Tables~\ref{tab:iam_related_work_prostate_desam_setup} and \ref{tab:in_domain_comparison} \\
& \textbf{HK (F)}~\cite{litjens2014evaluation} & as in DeSAM~\cite{gao2024desam} & 12 & 10 (90\%) & -- & 2 (10\%) & 12 (100\%) & Tables~\ref{tab:iam_related_work_prostate_desam_setup} and \ref{tab:in_domain_comparison} \\

\hline
\end{tabular}
}
\caption
[
Dataset overview.
]
{
Overview of the employed datasets and experimental setups.
The columns indicate: the dataset name and its original publication, the paper defining the employed setup, the total number of publicly available labeled samples, the respective number of samples in the train, validation, in-domain test and out-of-domain test set, as well as a list of tables from this manuscript in which the respective dataset and setup is employed.
$^{\dagger}$ Excluding cases with missing labels (CT: 5, MR: 1).
$^{\ddagger}$ Held out challenge test set with encrypted labels which requires using the evaluation script provided by the challenge organizers limiting evaluation.
$^{\mathsection}$ Excluding overall 9 cases, where we encountered issues with the orientation information.
}
\label{tab:dataset_table}
\end{table*}

%% file: tables/thale.t2.tex
\begin{table*}[t]
\centering
\footnotesize
\begin{tabular}{| l | c c c c | c | c c c c | c |}
\hline

\multicolumn{1}{|c}{\cellcolor{lightgray} \textbf{Abdominal}} & \multicolumn{5}{c}{\cellcolor{lightgray} \textbf{CT-1 $\rightarrow$ MR-1}} & \multicolumn{5}{c|}{\cellcolor{lightgray} \textbf{MR-1 $\rightarrow$ CT-1}} \\

\hline

\multicolumn{1}{|c|}{\multirow{2}{*}{\textbf{Method}}} & \multicolumn{5}{c|}{DSC (\%) $\uparrow$} & \multicolumn{5}{c|}{DSC (\%) $\uparrow$} \\

\cline{2-11}

& Liver & R-Kid. & L-Kid. & Spleen & Mean & Liver & R-Kid. & L-Kid. & Spleen & Mean \\

\hline

RSC~\cite{Huang2020-lc}\papercsdg & 76.40 & 75.79 & 76.60 & 67.56 & 74.09 & 88.10 & 46.60 & 75.94 & 53.61 & 66.07 \\
MixStyle~\cite{Zhou2021-ie}\papercsdg & 77.63 & 78.41 & 78.03 & 77.12 & 77.80 & 86.66 & 48.26 & 65.20 & 55.68 & 63.95 \\
AdvBias~\cite{Chen2020-xe}\papercsdg & 78.54 & 81.70 & 80.69 & 79.73 & 80.17 & 87.63 & 52.48 & 68.28 & 50.95 & 64.84 \\
SAMed~\cite{zhang2023customized}\paperdcon & -- & -- & -- & -- & -- & 84.86 & 56.04 & 67.71 & 66.47 & 68.77 \\
SAMUS~\cite{lin2024beyond}\paperdcon & -- & -- & -- & -- & -- & 84.73 & 61.56 & 71.39 & 64.25 & 70.48 \\
Cutout~\cite{DeVries2017-ad}\papercsdg & 79.80 & 82.32 & 82.14 & 76.24 & 80.12 & 86.99 & 63.66 & 73.74 & 57.60 & 70.50 \\
RandConv~\cite{Xu2020-vo}\papercsdg & 73.63 & 79.69 & 85.89 & 83.43 & 80.66 & 84.14 & 76.81 & 77.99 & 67.32 & 76.56 \\
ISW~\cite{choi2021robustnet}\paperirlsg & 84.83 & 82.60 & 83.44 & 81.06 & 82.98 & 83.23 & 68.88 & 76.47 & 70.06 & 74.66 \\
ATSRL~\cite{yang2021adversarial}\paperirlsg & 85.00 & 84.77 & 81.30 & 79.84 & 82.73 & 80.48 & 72.56 & 78.50 & 69.84 & 75.35 \\
BigAug~\cite{zhang2020generalizing}\paperirlsg & 85.51 & 84.76 & 82.09 & 81.16 & 83.38 & 85.17 & 73.35 & 77.40 & 73.92 & 77.46 \\
PCSDG~\cite{jiang2025structure}\paperarfu & 87.02 & 83.08 & 84.42 & 82.70 & 84.31 & 84.06 & 73.86 & 76.10 & 72.89 & 76.48 \\
ADS~\cite{xu2022adversarial}\paperirlsg & 85.24 & \textbf{89.87} & 86.92 & 81.69 & 85.93 & 83.06 & 76.85 & 79.44 & 73.05 & 78.10 \\
AugSeg~\cite{zhao2023augmentation}\paperdcon & -- & -- & -- & -- & -- & 89.19 & 80.38 & 80.38 & 69.97 & 79.98 \\
IRLSG~\cite{niu2024irlsg}\paperirlsg & 87.80 & 88.97 & 87.22 & 85.45 & 87.36 & 87.11 & 79.66 & 81.58 & 75.56 & 80.98 \\
ARFU~\cite{ren2025anatomically}\paperarfu & 87.56 & 88.54 & \underline{88.01} & \underline{89.49} & 88.40 & 87.61 & 81.56 & 81.92 & 80.14 & 82.81 \\
DCON~\cite{wang2025hybrid}\paperdcon & -- & -- & -- & -- & -- & \underline{89.98} & \underline{82.28} & \underline{82.95} & \underline{82.60} & \underline{84.45} \\
CSDG~\cite{Ouyang2023-xf}\papercsdg & 86.62 & 87.48 & 86.88 & 84.27 & 86.31 & 85.62 & 80.02 & 80.42 & 75.56 & 80.40 \\
SLAug~\cite{Su2023-du}\paperslaug & \textbf{90.08} & \underline{89.23} & 87.54 & 87.67 & \underline{88.63} & 89.26 & 80.98 & 82.05 & 79.93 & 83.05 \\

\hline

\multirow{2}{*}{\textbf{SRCSM}}
& \underline{89.88} & 88.46 & \textbf{88.31} & \textbf{91.32} & \textbf{89.49} & \textbf{92.37} & \textbf{84.83} & \textbf{84.71} & \textbf{88.53} & \textbf{87.61} \\
& $\pm$ 5.58 & $\pm$ 3.30 & $\pm$ 3.60 & $\pm$ 5.17 & $\pm$ 2.96 & $\pm$ 4.24 & $\pm$ 14.99 & $\pm$ 12.20 & $\pm$ 14.46 & $\pm$ 8.22 \\

\hline
\end{tabular}
\caption
[
Comparison to related work on abdominal data using the CSDG setup.
]
{
Comparison to related work on abdominal data using the CSDG setup.
Scores represent the DSC in \% per label as well as their mean.
Differently to related work, we also provide standard deviations for our method.
Scores for related work are obtained from:
\cite{Ouyang2023-xf}~(indicated by \papercsdg),
\cite{niu2024irlsg}~(\paperirlsg),
\cite{Su2023-du}~(\paperslaug),
\cite{wang2025hybrid}~(\paperdcon) and
\cite{ren2025anatomically}~(\paperarfu).
Number of images in the test sets: MR-1:~20, CT-1:~30.
}

\label{tab:related_work_abdominal_csdg_setup}
\end{table*}

%% file: tables/thale.t3.tex
\begin{table*}[t]
\centering
\footnotesize
\begin{tabular}{| l | l | c c | c c | c c | c c |}
\hline

\multicolumn{10}{|c|}{\cellcolor{lightgray} \textbf{Abdominal:} Confusion Matrix of Individual Source-Target Combinations} \\

\hline

\multicolumn{1}{|c|}{\multirow{1}{*}{}} & \multicolumn{1}{c|}{\multirow{4}{*}{\textbf{Method}}} & \multicolumn{8}{c|}{\textit{Target}} \\

\cline{3-10}

\multicolumn{1}{|c|}{\multirow{1}{*}{}} & & \multicolumn{2}{c|}{$\rightarrow$ \textbf{CT-1}} & \multicolumn{2}{c|}{$\rightarrow$ \textbf{CT-2}} & \multicolumn{2}{c|}{$\rightarrow$ \textbf{MR-1}} & \multicolumn{2}{c|}{$\rightarrow$ \textbf{MR-2}} \\

\multicolumn{1}{|c|}{\multirow{1}{*}{}} & & DSC & ASSD & DSC & ASSD & DSC & ASSD & DSC & ASSD \\
\multicolumn{1}{|c|}{\multirow{1}{*}{\textit{Source}}} & & (\%) $\uparrow$ & (mm) $\downarrow$ & (\%) $\uparrow$ & (mm) $\downarrow$ & (\%) $\uparrow$ & (mm) $\downarrow$ & (\%) $\uparrow$ & (mm) $\downarrow$ \\

\hline

\multirow{3}{*}{\textbf{CT-1}} 
& \multicolumn{1}{l|}{CSDG~\cite{Ouyang2023-xf}$^{\dagger}$} & -- & -- & 90.8 $\pm$ 6.1 & \underline{1.9} $\pm$ 2.1 & 85.7 $\pm$ 4.7 & 3.9 $\pm$ 1.4 & 88.7 $\pm$ 4.9 & 2.0 $\pm$ 1.1 \\
& \multicolumn{1}{l|}{SLAug~\cite{Su2023-du}$^{\dagger}$} & -- & -- & \underline{91.1} $\pm$ 5.7 & \underline{1.9} $\pm$ 2.2 & \underline{87.7} $\pm$ 3.6 & \underline{2.8} $\pm$ 1.2 & \underline{90.1} $\pm$ 4.5 & \underline{1.8} $\pm$ 1.0 \\
& \multicolumn{1}{l|}{\textbf{SRCSM}} & -- & -- & \textbf{93.0} $\pm$ 5.1 & \textbf{1.4} $\pm$ 2.5 & \textbf{89.5} $\pm$ 3.0 & \textbf{1.8} $\pm$ 0.8 & \textbf{92.9} $\pm$ 2.7 & \textbf{1.2} $\pm$ 0.6 \\

\hline

\multirow{3}{*}{\textbf{CT-2}} 
& \multicolumn{1}{l|}{CSDG~\cite{Ouyang2023-xf}$^{\dagger}$} & \underline{92.3} $\pm$ 6.3 & \underline{1.9} $\pm$ 2.5 & -- & -- & 86.9 $\pm$ 6.3 & 3.9 $\pm$ 2.0 & \underline{92.7} $\pm$ 3.7 & \underline{1.1} $\pm$ 0.9 \\
& \multicolumn{1}{l|}{SLAug~\cite{Su2023-du}$^{\dagger}$} & \textbf{92.9} $\pm$ 6.0 & \textbf{1.6} $\pm$ 2.1 & -- & -- & \underline{87.0} $\pm$ 6.4 & \underline{3.7} $\pm$ 2.5 & \underline{92.7} $\pm$ 3.6 & 1.3 $\pm$ 0.8 \\
& \multicolumn{1}{l|}{\textbf{SRCSM}} & 91.2 $\pm$ 7.1 & \textbf{1.6} $\pm$ 2.2 & -- & -- & \textbf{89.5} $\pm$ 5.3 & \textbf{1.9} $\pm$ 1.7 & \textbf{93.9} $\pm$ 1.9 & \textbf{1.0} $\pm$ 0.4 \\

\hline

\multirow{3}{*}{\textbf{MR-1}} 
& \multicolumn{1}{l|}{CSDG~\cite{Ouyang2023-xf}$^{\dagger}$} & 80.1 $\pm$ 9.6 & 6.9 $\pm$ 3.9 & 75.3 $\pm$ 13.3 & 5.5 $\pm$ 3.2 & -- & -- & 84.9 $\pm$ 8.1 & 2.8 $\pm$ 1.6 \\
& \multicolumn{1}{l|}{SLAug~\cite{Su2023-du}$^{\dagger}$} & \underline{82.5} $\pm$ 8.6 & \underline{4.4} $\pm$ 2.9 & \underline{78.3} $\pm$ 12.3 & \textbf{4.5} $\pm$ 3.2 & -- & -- & \underline{86.8} $\pm$ 5.9 & \underline{2.5} $\pm$ 1.3 \\
& \multicolumn{1}{l|}{\textbf{SRCSM}} & \textbf{87.6} $\pm$ 8.2 & \textbf{3.7} $\pm$ 3.1 & \textbf{85.6} $\pm$ 10.0 & \underline{5.1} $\pm$ 5.1 & -- & -- & \textbf{92.1} $\pm$ 2.6 & \textbf{2.0} $\pm$ 0.9 \\

\hline

\multirow{3}{*}{\textbf{MR-2}} 
& \multicolumn{1}{l|}{CSDG~\cite{Ouyang2023-xf}$^{\dagger}$} & \underline{88.8} $\pm$ 9.0 & \underline{3.0} $\pm$ 3.2 & \underline{89.9} $\pm$ 7.3 & \underline{1.9} $\pm$ 2.5 & \underline{90.0} $\pm$ 2.8 & \underline{1.9} $\pm$ 0.8 & -- & -- \\
& \multicolumn{1}{l|}{SLAug~\cite{Su2023-du}$^{\dagger}$} & 80.9 $\pm$ 10.0 & 4.8 $\pm$ 4.1 & 77.5 $\pm$ 13.2 & 4.5 $\pm$ 3.5 & 85.5 $\pm$ 4.9 & 3.7 $\pm$ 1.6 & -- & -- \\
& \multicolumn{1}{l|}{\textbf{SRCSM}} & \textbf{91.1} $\pm$ 6.7 & \textbf{1.9} $\pm$ 2.0 & \textbf{92.4} $\pm$ 5.9 & \textbf{1.6} $\pm$ 2.7 & \textbf{91.9} $\pm$ 2.6 & \textbf{1.5} $\pm$ 0.8 & -- & -- \\

\hline

\end{tabular}
\caption{
Detailed quantitative results for all source-target combinations on abdominal data of our SRCSM method compared to SLAug and CSDG in the CSDG setup.
Presented are the mean and standard deviation of the DSC in \% as well as the ASSD in mm.
$^{\dagger}$~indicates scores obtained using the original authors code.
Number of images in the test sets: CT-1:~30, CT-2:~295, MR-1:~20, MR-2:~59.
}
\label{tab:abdominal_detailed}
\end{table*}

%% file: our_paper/discussion.tex
\section{Results and Discussion}

\input{tables/thale.t4}

\input{tables/thale.t5}

\input{tables/thale.t6}

\subsection{Quantitative Comparison to Related Work}
\label{sec:quantitative_comparison_related_work}

We provide a large-scale comparison to related work with a focus on recent single-source \gls{dg} strategies in Tables~\ref{tab:related_work_abdominal_csdg_setup}-\ref{tab:iam_related_work_prostate_desam_setup}.
Beyond this, we also provide comparisons to \gls{da} and \gls{tta} approaches in Tables~\ref{tab:related_work_mmwhs} and~\ref{tab:iam_related_work_prostate_hsd_setup}.

\subsubsection*{Cross-modality Abdominal Organ Segmentation}

We present the performance in abdominal organ segmentation in Tables~\ref{tab:related_work_abdominal_csdg_setup}-\ref{tab:related_work_abdominal_desam_setup}.
As shown in Table~\ref{tab:related_work_abdominal_csdg_setup}, \gls{srcsm} outperforms a large variety of single-source \gls{dg} baselines in both cross-modality settings.
On average, \gls{srcsm} achieved an improvement of $+0.86$\% \gls{dsc} when generalizing from \gls{ct}-1 to \gls{mr}-1 compared to the second best baseline.
In the opposite direction, from \gls{mr}-1 to \gls{ct}-1, \gls{srcsm} outperforms all baselines by at least $3.16$\% \gls{dsc}.
\gls{srcsm}'s \gls{ci} for the mean \gls{dsc} in the \gls{ct}-1 $\rightarrow$ \gls{mr}-1 setting is $[88.11,90.87]$; only two related works achieve scores that lie within this range.
In the opposite direction, i.e., when generalizing from \gls{mr}-1 to \gls{ct}-1, the \gls{ci} of our method for the mean \gls{dsc} is $[84.55,90.67]$ with no related work score falling within this range.
Most notably, \gls{srcsm} achieved the largest performance improvements when segmenting the spleen, with $+1.83$\% \gls{dsc} when generalizing from \gls{mr}-1 to \gls{ct}-1 and $+5.93$\% \gls{dsc} in the opposite direction.

\input{tables/thale.t7}

\input{tables/thale.t8}

We present a more detailed analysis of abdominal organ segmentation by adding an additional \gls{ct} and \gls{mr} dataset in Table~\ref{tab:abdominal_detailed}, which we respectively call \gls{ct}-2 and \gls{mr}-2 as described in Section~\ref{sec:datasets}.
With this, we increase the number of possible abdominal source-target combinations from 2 to 12, for which we selected two baselines: SLAug~\cite{Su2023-du} as a method that also augments data based on class labels and CSDG~\cite{Ouyang2023-xf} as a popular baseline that also relies on random convolutions.
Overall, \gls{srcsm} respectively achieves the best \gls{dsc} and the best \gls{asd} result in 11 out of 12 source-target combinations, while performing on par with related work in the remaining ones.
On average, \gls{srcsm} outperforms the respective second best baseline by $+2.7$\% \gls{dsc} and $-0.4\text{ mm}$ \gls{asd}.

While Table~\ref{tab:related_work_abdominal_csdg_setup} already includes a comparison to some SAM-based methods~\cite{zhang2023customized,lin2024beyond}, we performed additional experiments in abdominal organ segmentation in the same setup as DeSAM~\cite{gao2024desam} (see Section~\ref{sec:datasets}) to allow additional comparisons to~\cite{zhang2023customized,gao2024desam,ma2024segment}.
The results are presented in Table~\ref{tab:related_work_abdominal_desam_setup}.
\gls{srcsm} outperforms all baselines in both source-target domain pairs by a large margin.
The \glspl{ci} of \gls{srcsm} computed for the \gls{dsc} respectively are $[89.35,91.73]$ for \gls{ct}-1 $\rightarrow$ \gls{mr}-1 and $[87.07,91.95]$ for \gls{mr}-1 $\rightarrow$ \gls{ct}-1.
Closer inspection shows that no related work score falls within the respective \gls{ci}.
Notably, in the \gls{ct}-1 $\rightarrow$ \gls{mr}-1 setting, a large gap remains between the best performing related work score and the lower bound of \gls{srcsm}'s \gls{ci}.

\subsubsection*{Cross-modality Whole-Heart Segmentation}

When segmenting the heart in the four-label setting, in Table~\ref{tab:related_work_mmwhs}, our method achieves the best results with a gap of $+5.1\%$ \gls{dsc} and $-0.5\text{ mm}$ \gls{asd} over other single-source \gls{dg} methods.
Most notably, \gls{srcsm} even outperforms recent \gls{da} approaches by at least $4.7$\% \gls{dsc}, despite not demanding simultaneous access to source and target domain data during training.
In the more detailed setting with seven labels, in Table~\ref{tab:related_work_mmwhs_7labels}, our method also achieves great improvements over SLAug and CSDG: $+6.4$\% \gls{dsc} when generalizing from \gls{mr} to \gls{ct} and $+7.2$\% \gls{dsc} in the other direction.

\subsubsection*{Cross-center Prostate Segmentation}

\input{figures/thale6}

The results of prostate segmentation, presented in
Table~\ref{tab:iam_related_work_prostate_hsd_setup}, show that \gls{srcsm} outperforms related work in 4 out of 5 target domains by up to $+4.7$\% \gls{dsc}, while achieving a similar performance to the most competitive baseline in the fifth one.
When the scores are averaged over all target domains, \gls{srcsm} achieves improvements of $+1.8$\% \gls{dsc} over related \gls{dg} approaches and $+8.5$\% \gls{dsc} over \gls{tta} methods.
In terms of average \gls{hd}, \gls{srcsm} performs in-line with HSD in the transformer variant, while outperforming all other methods by a large margin.
These are notable improvements especially when considering that \gls{tta} employs target domain data to adapt model parameters.
We also computed the \glspl{ci} from the \gls{dsc} scores achieved on each domain by \gls{srcsm} which are: B:~$[90.02,91.58]$, C:~$[88.02,91.38]$, D:~$[88.46,92.54]$, E:~$[83.56,88.84]$ and F:~$[89.35,92.25]$.
HSD in the transformer variant is the only related work method that achieved a score that falls within the \gls{ci} of \gls{srcsm} for two out of five domains, namely, when generalizing to prostate E and prostate F.
The results in Table~\ref{tab:iam_related_work_prostate_desam_setup}, i.e., when training on respectively one prostate domain and evaluating on the union of the other domains, show that \gls{srcsm} substantially outperformed all other methods, achieving an improvement of at least $+9.13$\% \gls{dsc} over other methods on average.
Notably, in the setting which is most challenging for related work, i.e., prostate C~$\rightarrow$~Rest, \gls{srcsm} achieved a substantial improvement of $+17.66\%$ over the second best method.
Closer inspection of \gls{srcsm}'s \glspl{ci} in terms of \gls{dsc} for each source domain shows that no score achieved by related work falls into the respective \gls{ci} of \gls{srcsm} and, moreover, that large gaps remain between the lower bound of the \gls{ci} and the respective best score achieved by related work.
The \glspl{ci} of \gls{srcsm} are: A:~$[89.03,90.49]$, B:~$[88.16,90.46]$, C:~$[81.23,87.39]$, E:~$[89.68,90.88]$, D:~$[87.59,89.23]$, F:~$[84.97,88.73]$.

\subsubsection*{Statistical Analysis of Multiple Dataset Generalization}

Based on results in Tables~\ref{tab:abdominal_detailed} and~\ref{tab:related_work_mmwhs_7labels}, we can compare statistical significance of the performance differences of our proposed \gls{srcsm} with CSDG and SLAug on 14 different datasets, as described in Section~\ref{sec:evaluation_and_statistical_comparison}. 
Autorank~\cite{Herbold2020-ea} assessed that the population's (algorithm's) performances deviated from a normal distribution and therefore performed a non-parametric ranking-based Friedman test to test if there were any differences between populations. 
Since at least one difference was found, Autorank selected a post-hoc Nemenyi test to assess differences between pairs of algorithms. 
The results in Figure~\ref{fig:ranking} show that for both metrics, \gls{dsc} and \gls{asd}, \gls{srcsm} outperforms SLAug and CSDG at a significance level of $p = 0.05$, whereas SLAug and CSDG remain within \gls{cd} to one another.

\input{tables/thale.t9}

\input{figures/thale7}

\subsection{Comparison to In-Domain Performance}
\label{sec:quantitative_comparison_to_indomain}

We compare the performance of \gls{srcsm} to that of a model trained on the target domain, in Tables~\ref{tab:related_work_mmwhs} and~\ref{tab:in_domain_comparison}, respectively.
Importantly, all scores in Table~\ref{tab:in_domain_comparison} have been computed on held-out validation respectively test sets to allow for a fair comparison of \gls{srcsm} to the in-domain baseline.
This is necessary since most single-source \gls{dg} setups use all available target domain data in their evaluation and therefore leave no training data for an in-domain baseline, thus preventing a direct comparison.

The results in Table~\ref{tab:related_work_mmwhs} show that \gls{srcsm} closes in on the in-domain baseline, effectively reducing the remaining domain gap to $1.5$\% \gls{dsc} and $0.3\text{ mm}$ \gls{asd}.
For the best performing related work method, a considerably larger domain gap of $6.2$\% \gls{dsc} and $0.8\text{ mm}$ \gls{asd} remains.

Moreover, we present a thorough comparison to an in-domain baseline in Table~\ref{tab:in_domain_comparison}, which includes abdominal, seven label whole-heart and prostate domains.
For several source-target combinations, \gls{srcsm} performs in-line with or even slightly exceeds the in-domain baseline, while closing in on its results in most remaining settings.
Notably, when averaging over the 16 domains in Table~\ref{tab:in_domain_comparison}, \gls{srcsm} reduces the remaining domain gap to a respectable $2.8$\% \gls{dsc}.
Larger performance gaps to the in-domain baseline only occurred in two out of 16 source-target combinations, which we analyze below.

Generalization in the Each~$\rightarrow$~C setting results in a relatively large disparity between the performance of \gls{srcsm} and the in-domain baseline, amounting to $12.0$\% \gls{dsc}.
We suspect that this comparatively large performance gap of Each~$\rightarrow$~C, as compared to the other prostate settings, stems from a larger domain gap of prostate C to the other domains.
Besides differences in coil usage and the hardware used for image acquisition, another difference between domains arises from the health status of the individual subjects~\cite{litjens2014evaluation,liu2020ms}.
Specifically, prostate C has a very high incidence rate of prostate cancer~\cite{liu2020ms}, which additionally contributes to the domain gap of prostate C.
Further evidence on a larger domain gap of prostate C can be found in Table~\ref{tab:iam_related_work_prostate_desam_setup}, where the prostate C~$\rightarrow$~Rest setting also underperforms compared to other settings for related work methods.

\input{tables/thale.t10}

Another large performance gap in Table~\ref{tab:in_domain_comparison} is observed for the abdominal \gls{mr}-1~$\rightarrow$~\gls{ct}-2 setting, where \gls{srcsm} attains a performance that is $9.5$\% lower as compared to the in-domain baseline.
Interestingly, generalizing from the other abdominal \gls{mr} domain, i.e., \gls{mr}-2~$\rightarrow$~\gls{ct}-2, results in a much lower performance difference to the in-domain score with only $2.7$\% as compared to the $9.5$\% \gls{dsc} of \gls{mr}-1~$\rightarrow$~\gls{ct}-2.
Upon closer inspection of the individual abdominal datasets, we discovered that the ground truth segmentations of \gls{mr}-1 are inconsistent with the segmentations of the other abdominal datasets.
Specifically, \gls{mr}-1 is the only domain that also includes the renal pelvis in the ground truth segmentation of the left and right kidney, which makes settings that involve \gls{mr}-1 inherently more challenging.
Further, closer inspection of the related work comparison in Table~\ref{tab:abdominal_detailed} confirms that other methods also underperform on the \gls{mr}-1~$\rightarrow$~\gls{ct}-2 setting when compared to their respective performance on the other source-target combinations.
This indicates that \gls{mr}-1~$\rightarrow$~\gls{ct}-2 is a particularly difficult abdominal cross-modality setting which is partially caused by the difference in the ground truth segmentation of the source and target domain.

\subsection{Qualitative Comparison to Related Work}
\label{sec:qualitative_comparison_related_work}

The qualitative results, shown in Fig.~\ref{fig:qualitative}, demonstrate that \gls{srcsm} produced very accurate segmentation masks.
The observed errors did not affect larger parts of anatomical structures:
A slight undersegmentation of the spleen (yellow) can be observed when adapting from \gls{ct}-1 to \gls{mr}-1 (row~2, col~3), which however, did not occur when adapting from \gls{ct}-2 to \gls{mr}-1 (row~2, col~6).
When adapting on the cardiac whole-heart dataset from \gls{ct} to \gls{mr}, \gls{srcsm} slightly oversegmented the right ventricle (green) (row~3, col~7).
In either case, the segmentation masks produced by \gls{srcsm} are more plausible than those yielded by the baselines.
Interestingly, when adapting from abdominal \gls{mr}-1 to \gls{ct}-1 (row~1, col~3), it can be observed that both, the left (green) and right kidney (blue) were labeled as such that the renal pelvis is respectively included in the predicted segmentation.
This is caused by a mismatch in the ground truth segmentation of \gls{mr}-1 when compared to the other domains.

\subsection{Whole-Heart to Cine Experiments}
\label{sec:whole_heart_to_cine}

So far, the evaluation in this work focused on domain gaps introduced by differences in intensity, contrast and noise.
To extend the evaluation to geometric domain shifts, we additionally assessed generalization from whole-heart \gls{ct} and \gls{mr} to cardiac cine \gls{mr} data.
In this challenging setting, additional sources of domain shift arise, including (1) large slice gaps in the out-of-plane dimension, (2) potential misalignment of neighboring slices due to motion, and (3) morphological changes from diastolic whole-heart to the systolic phase of cine \gls{mr} data.
To our knowledge, this is the first work that evaluates generalization from whole-heart to cine \gls{mr} data.

The results in Table~\ref{tab:whs_to_cine} show that on average, \gls{srcsm} leads to respectable improvements over a baseline that only employs conventional data augmentation (CDA) when generalizing from whole-heart \gls{mr} to cine \gls{mr}.
Most notably, \gls{srcsm} greatly outperforms the baseline in the whole-heart \gls{ct} to cine \gls{mr} setting, where the baseline fails completely, while \gls{srcsm} fares as well as when adapting from whole-heart \gls{mr} data.
Differently to the baseline, \gls{srcsm} achieves a similar performance for diastolic and the more challenging systolic scans, which emphasizes the potential of strong intensity augmentation techniques even when large morphological differences are expected.

\input{tables/thale.t11}

\subsection{Ablation Study}
\label{sec:ablation_study}

The ablation study in Table~\ref{tab:ablation} demonstrates the contribution of the individual components of \gls{srcsm} to the overall performance.
By design, the two complementary components address different potential sources of domain gap, which is confirmed by the fact that their combination systematically improves performance.
Interestingly, in the abdominal \gls{mr}-1 to \gls{ct}-1 setting, \gls{srcsm} using the binary \gls{src} implementation achieved slightly better scores then \gls{srcsm} as proposed, however, only by relatively small amounts of $0.4$\% \gls{dsc} and $0.1 \text{ mm}$ \gls{asd}.
Closer inspection of the data shows that the physical resolution of \gls{mr}-1 averages to roughly $1.55 \text{ mm}$ in-plane and $8.75 \text{ mm}$ out-of-plane, which is much lower than the average physical resolution of \gls{ct}-1 with roughly $0.78 \text{ mm}$ in-plane and $3.78 \text{ mm}$ out-of-plane.
This results in the \gls{mr}-1 images being inherently less clear and appearing more blurry in comparison to \gls{ct}-1 images.
Moreover, as described in Section~\ref{sec:quantitative_comparison_to_indomain}, we identified a disparity in the kidney ground truth segmentation of \gls{mr}-1 as compared to the other abdominal domains.
These factors may have contributed to the small performance gap in favor of the binary \gls{src} implementation as observed in the abdominal \gls{mr}-1 to \gls{ct}-1 setting.
Nevertheless, in all settings shown in Table~\ref{tab:ablation}, using our proposed \gls{src} and \gls{sm} results in the highest performance.

\input{figures/thale8}

\subsection{Analysis of Intermediate Features}

The t-SNE plots of our ablation study in Fig.~\ref{fig:tsne_all} show that our contributions improve the alignment of intermediate feature representations of points sampled from the source (in-domain) and target (out-of-domain) domain.
While the \gls{cda} baseline (row~1) successfully forms distinct clusters per label (colors) for in-domain points (row~1, col~1), it is unable to produce a similar clustering for out-of-domain points (row~1, col~2).
In addition, when superimposing in-domain and out-of-domain points, they are completely misaligned (row~1, col~3).
When employing only one of our contributions, \gls{src} (row~2) or \gls{sm} (row~3), the t-SNE plots show that both improve the formation of distinct clusters per label for out-of-domain points (row~2-3, col~2) as well as the alignment between in-domain and out-of-domain points (row~2-3, col~3).
Importantly, while \gls{src} succeeds in the formation of per label clusters for out-of-domain points (row~2, col~2), \gls{sm} results in a better alignment of in-domain and out-of-domain points (row~3, col~3).
The complementary nature of \gls{src} and \gls{sm} allows the proposed \gls{srcsm} method to benefit from combining both contributions (row~4).
Specifically, \gls{srcsm} results in the formation of clearly distinct clusters per label for out-of-domain points (row~4, col~2) as well as in the best alignment of in-domain and out-of-domain points (row~4, col~3) among all variants.
To complement the t-SNE plots, we also provide exemplary images in Fig.~\ref{fig:tsne_all} (col~4-5), where the ground truth segmentation for each label is visualized as a contour line, while model predictions are visualized as colored semi-transparent areas.
The predictions shown in these images have been produced by the same models shown in the t-SNE plots and underline the complementary domain generalization improvements achieved by our contributions. These results support our initial motivation that \gls{dg} should be addressed via a two-step approach, namely, by (1) expanding the space of the source domain's data distribution during training and by (2) aligning target domain data to the source domain.

\subsection{Qualitative Error Analysis}

Closer inspection on a case-by-case basis revealed that only very few outstanding failure cases exist across datasets.
We present examples from different source-target combinations in Fig.~\ref{fig:case_level_error_analysis}, which correspond to experiments in our ablation study to also allow assessing how our individual contributions influence the prediction.
In this figure, we present examples from the following settings:
abdominal \gls{mr}-1~$\rightarrow$~\gls{ct}-1 (row~1) and
\gls{ct}-1~$\rightarrow$~\gls{mr}-1 (row~2), as well as 
cardiac \gls{mr}~$\rightarrow$~\gls{ct} (row~3) and
\gls{ct}~$\rightarrow$~\gls{mr} (row~4).
We show the image (col~1), the image overlaid with the prediction produced by the \gls{cda} baseline (col~2), \gls{cda}+\gls{src} (col~3), \gls{cda}+\gls{sm} (col~4), the proposed \gls{srcsm} method which employs \gls{cda}+\gls{src}+\gls{sm} (col~5), as well as the image overlaid with the ground truth annotation (col~6).
Interestingly, error cases mostly comprise images that -- to our knowledge -- show pathologies which affect organ morphology.
In particular, the examples appear to include a subject with kidney disease (row~1, green and blue), a subject with an enlarged spleen (row~2, yellow), and a subject with right ventricle dilation (row~4, green).
For the shown cases, we observed that both \gls{src} (col~3) and \gls{sm} (col~4) result in improvements over the \gls{cda} baseline (col~2) with \gls{src} generally leading to larger improvements.
Additional improvements are achieved by combining our contributions and employing the proposed \gls{srcsm} method (col~5).
Note that the abdominal \gls{ct}-1~$\rightarrow$~\gls{mr}-1 example (row~2) is also shown in our qualitative comparison to related work in Fig.~\ref{fig:qualitative} (row~2).
This allows an additional comparison to SLAug (row~2, col~4) and CSDG (row~2, col~5) which result in worse or similar prediction errors, respectively.
Even though some predictions of \gls{srcsm} remain partially erroneous, we want to emphasize that \gls{srcsm} is not designed to address morphology-based domain gaps and therefore assumes that organ morphology between the source and target domain is comparable.
As shown in Fig.~\ref{fig:qualitative} (row~2, col~6), \gls{srcsm} resolves the described undersegmentation of the enlarged spleen (yellow) when generalizing from the \gls{ct}-2 dataset, which is significantly larger than \gls{ct}-1.

\input{figures/thale9}

\subsection{Limitations}
\label{sec:limitations}

Like previous methods, \gls{srcsm} assumes that images of the source and target domains show the same anatomy and offer roughly the same field of view.
In particular, \gls{sm} might be sensitive to drastic changes in the field of view for which further evaluation is necessary.
However, previous localization of the region of interest, e.g., with the method in~\cite{payer2019integrating}, provides a straight-forward solution to alleviate this issue.
\gls{sm} might also be sensitive to imaging artifacts that alter the intensity distribution of an image, similarly to many common normalization techniques.
While further analysis is left for future work, the whole-heart to cine experiments provide additional insight on the generality of our method by assessing the performance in the presence of domain gaps our method was not explicitly designed to bridge.
These domain gaps include morphological differences, potential slice misalignment of neighboring cine slices, differences in label definition and changes in field of view.
While our method achieved good results in these experiments, we acknowledge that a performance gap remains when compared to the in-domain results.
Further investigation is necessary to assess how much additional improvement can be achieved when these additional sources of domain gap are addressed, e.g., by registering cine slices and carefully assessing and potentially aligning the label definition.

Even though we perform a comprehensive evaluation for cross-modality \gls{dg}, some limitations related to the evaluation remain.
While we employ strong spatial augmentation to account for possible morphological differences between source and target domain, the augmentation techniques are limited by their respective predefined parameterization.
Further assessment of the performance in the presence of significant morphological differences is left for future work.
Our evaluation on the systolic cine data already provides some evidence on the generality of our method under morphological changes, however, additional experiments on cases of rare disease types or using, e.g., abdominal scans that contain organs with tumors of different sizes and at different locations, would be interesting.
Further, in this work, we focus on the challenging single-source cross-modality setting between \gls{mr}- and \gls{ct}-based imaging modalities and therefore we specifically designed \gls{sm} with grayscale images in mind.
While an extension to multi-channel data like RGB images in histopathology is technically possible, for example, by applying \gls{sm} to each channel independently, this goes beyond the scope of this work and is left for future extension.

%% file: tables/thale.t4.tex
\begin{table}[t]
\centering
\footnotesize
\begin{tabular}{| l | c | c |}
\hline

\multicolumn{1}{|c}{\cellcolor{lightgray} \textbf{Abdominal}} & \multicolumn{1}{c}{\cellcolor{lightgray} \textbf{CT-1 $\rightarrow$ MR-1}} & \multicolumn{1}{c|}{\cellcolor{lightgray} \textbf{MR-1 $\rightarrow$ CT-1}} \\

\hline

\multicolumn{1}{|c|}{\multirow{1}{*}{\textbf{Method}}} & \multicolumn{1}{c|}{DSC (\%) $\uparrow$} & \multicolumn{1}{c|}{DSC (\%) $\uparrow$} \\

\hline

MaxStyle~\cite{chen2022maxstyle}\paperdesam & 76.93 & 82.92 \\
CSDG~\cite{Ouyang2023-xf}\paperdesam & 77.54 & 83.57 \\

MedSAM~\cite{ma2024segment}\paperdesam & 72.10 & 80.64 \\
SAMed~\cite{zhang2023customized}\paperdesam & 70.35 & 77.21 \\
DeSAM-B~\cite{gao2024desam}\paperdesam & 79.57 & 84.87 \\
DeSAM-P~\cite{gao2024desam}\paperdesam & \underline{80.05} & \underline{86.68} \\

\hline

\textbf{SRCSM} & \textbf{90.54} $\pm$ 2.55 & \textbf{89.51} $\pm$ 6.54 \\

\hline
\end{tabular}
\caption
[
Comparison to related work on abdominal data using the DeSAM setup.
]
{
Comparison to related work on abdominal data using the DeSAM setup.
Scores represent the mean DSC in~\%.
Differently to related work, we also report standard deviations.
Scores for related work are obtained from: \cite{gao2024desam}~(\paperdesam).
Number of images in the test sets: MR-1:~20, CT-1:~30.
}
\label{tab:related_work_abdominal_desam_setup}
\end{table}

%% file: tables/thale.t5.tex
\begin{table*}[t]
\centering
\footnotesize
\begin{tabular}{| c l | c c c c | c | c |}
\hline

\multicolumn{2}{|c}{\cellcolor{lightgray} \textbf{Cardiac:} 4 Labels} & \multicolumn{6}{c|}{\cellcolor{lightgray} \textbf{MR $\rightarrow$ CT}} \\
\hline

\multicolumn{2}{|c|}{\multirow{3}{*}{\textbf{Method}}} & \multicolumn{5}{c|}{\multirow{2}{*}{DSC (\%) $\uparrow$}} & \multicolumn{1}{c|}{ASSD} \\
& & \multicolumn{5}{c|}{} & \multicolumn{1}{c|}{(vox) $\downarrow$} \\

\cline{3-8}
& & AA & LA & LV & MYO & Mean & Mean \\
\hline

& In-Domain & 94.2 $\pm$ 6.3 & 92.2 $\pm$ 2.6 & 92.3 $\pm$ 4.5 & 90.3 $\pm$ 1.9 & 92.3 $\pm$ 2.5 & 1.2 $\pm$ 0.4 \\

\hline

\multirow{6}{*}{\textit{\rotatebox{90}{DA}}}
& CycleGAN~\cite{Zhu2017-sj}\paperadami & 73.8 & 75.7 & 52.3 & 28.7 & 57.6 & 10.8 \\
& SIFA~\cite{Chen2020-kh}\paperadami & 81.3 & 79.5 & 73.8 & 61.6 & 74.1 & 7.0 \\
& SynthMix~\cite{zhang2023synthmix}\papercthreer & 87.2 & 88.5 & 82.4 & 71.8 & 82.5 & -- \\
& CUDA~\cite{Du2022-xe}\papercuda & 87.2 & 88.5 & 83.0 & 72.8 & 82.9 & 5.5 \\
& MPSCL~\cite{Liu2022-zd}\papermpscl & 90.3 & 87.1 & 86.5 & 72.5 & 84.1 & 3.5 \\
& C3R~\cite{ding2025c3r}\papercthreer & 90.2 & 90.2 & \underline{86.6} & \underline{77.5} & \underline{86.1} & 3.3 \\

\hline

\multirow{4}{*}{\textit{\rotatebox{90}{TTA}}}
& Prior AdaEnt~\cite{Bateson2020-gl}\paperadami & 75.5 & 71.2 & 59.4 & 56.4 & 65.6 & 8.2 \\
& AdaMI~\cite{Bateson2022-bv}\paperadami & 83.1 & 78.2 & 74.5 & 66.8 & 75.7 & 5.6 \\
& Tent~\cite{Wang2021-ho}\paperttas & 55.4 & 33.4 & 63.0 & 41.1 & 48.2 & 11.2 \\
& TTAS$_\EuScript{RC}$~\cite{Bateson2022-gh}\paperttas & 85.1 & 82.6 & 79.3 & 73.2 & 80.0 & 5.3 \\

\hline

\multirow{3}{*}{\textit{\rotatebox{90}{DG}}}
& CSDG~\cite{Ouyang2023-xf}$^{\dagger}$ & \underline{95.4} $\pm$ 0.8 & 90.9 $\pm$ 1.3 & 82.0 $\pm$ 10.8 & 71.2 $\pm$ 5.1 & 84.9 $\pm$ 3.3 & \underline{2.0} $\pm$ 0.5 \\
& SLAug~\cite{Su2023-du}$^{\dagger}$ & \textbf{95.8} $\pm$ 0.4 & \underline{91.2} $\pm$ 1.3 & 82.0 $\pm$ 10.7 & 73.7 $\pm$ 4.5 & 85.7 $\pm$ 3.2 & 2.4 $\pm$ 0.5 \\

\cline{2-8}

& \textbf{SRCSM} & 94.0 $\pm$ 1.3 & \textbf{91.7} $\pm$ 2.3 & \textbf{90.4} $\pm$ 5.2 & \textbf{87.0} $\pm$ 1.3 & \textbf{90.8} $\pm$ 0.7 & \textbf{1.5} $\pm$ 0.3 \\

\hline

\end{tabular}
\caption
[
Comparison to related work on the whole-heart dataset using four labels.
]
{
Comparison to related work on the whole-heart dataset using four labels.
Shown are the DSC in \% per label as well as the mean and standard deviation ofnd ASSD in voxel.
Labelsare: ascending aorta (AA), left atrium (LA), left ventricle (LV) and myocardium (MYO).
Scores for related work are obtained from:
\cite{Bateson2022-bv}~(\paperadami),
\cite{ding2025c3r}~(\papercthreer),
\cite{Du2022-xe}~(\papercuda),
\cite{Liu2022-zd}~(\papermpscl), 
\cite{Bateson2022-gh}~(\paperttas).
Number of images in the test set:~4.
$^{\dagger}$~indicates scores obtained using the original authors code.
}

\label{tab:related_work_mmwhs}
\end{table*}

%% file: tables/thale.t6.tex
\begin{table}[htbp]
\centering
\footnotesize

\resizebox{\columnwidth}{!}{%
\begin{tabular}{| l | c c | c c |}
\hline

\multicolumn{1}{|c}{\cellcolor{lightgray} \textbf{Card.:} 7 Lab.} & \multicolumn{2}{c}{\cellcolor{lightgray} \textbf{CT $\rightarrow$ MR}} & \multicolumn{2}{c|}{\cellcolor{lightgray} \textbf{MR $\rightarrow$ CT}} \\

\hline

\multicolumn{1}{|c|}{\multirow{2}{*}{\textbf{Method}}} & DSC & ASSD & DSC & ASSD \\
& (\%) $\uparrow$ & (mm) $\downarrow$ & (\%) $\uparrow$ & (mm) $\downarrow$ \\

\hline

CSDG~\cite{Ouyang2023-xf}$^{\dagger}$ & 77.3 $\pm$ 9.5 & \underline{2.6} $\pm$ 1.4 & \underline{82.3} $\pm$ 3.4 & \underline{2.9} $\pm$ 1.0 \\
SLAug~\cite{Su2023-du}$^{\dagger}$ & \underline{78.5} $\pm$ 8.1 & 3.1 $\pm$ 1.6 & 82.2 $\pm$ 2.8 & 3.6 $\pm$ 1.3 \\

\hline

\textbf{SRCSM} & \textbf{85.7} $\pm$ 2.8 & \textbf{1.8} $\pm$ 0.6 & \textbf{88.7} $\pm$ 2.1 & \textbf{1.4} $\pm$ 0.4 \\

\hline

\end{tabular}
}
\caption{
Comparison to related work on the whole-heart dataset using seven labels.
Presented are the mean and standard deviation of the DSC in \% as well as the ASSD in mm.
Number of images in the test sets: MR:~20, CT:~20.
$^{\dagger}$~indicates scores obtained using the original authors code.
}
\label{tab:related_work_mmwhs_7labels}
\end{table}

%% file: tables/thale.t7.tex
\begin{table*}[t]
\centering
\footnotesize
\begin{tabular}{| c l | c c c c c | c | c |}
\hline

\multicolumn{2}{|c}{\cellcolor{lightgray} \textbf{Prostate}} & \multicolumn{7}{c|}{\cellcolor{lightgray} \textbf{A $\rightarrow$ $\doubleunderscore$}} \\

\hline

\multicolumn{2}{|c|}{\multirow{3}{*}{\textbf{Method}}} & \multicolumn{6}{c|}{\multirow{1}{*}{DSC (\%) $\uparrow$}} & \multicolumn{1}{c|}{\multirow{1}{*}{HD95 (mm) $\downarrow$}} \\

\cline{3-9}
& & \multirow{2}{*}{B} & \multirow{2}{*}{C} & \multirow{2}{*}{D} & \multirow{2}{*}{E} & \multirow{2}{*}{F} & Mean over & Mean over \\
& & & & & & & Domains & Domains \\

\hline

\multirow{4}{*}{\textit{\rotatebox{90}{TTA}}}
& TTT~\cite{sun2020test}\paperhsd & 83.5 $\pm$ 5.9 & 73.1 $\pm$ 17.5 & 75.3 $\pm$ 7.8 & 67.5 $\pm$ 11.1 & 81.5 $\pm$ 5.9 & 76.2 $\pm$ 6.5 & 19.8 $\pm$ 7.8 \\
& TTST~\cite{Karani2021-hh}\paperhsd & 86.0 $\pm$ 3.7 & 74.8 $\pm$ 10.5 & 81.0 $\pm$ 3.9 & 74.0 $\pm$ 8.4 & 80.9 $\pm$ 9.2 & 79.3 $\pm$ 5.0 & 18.3 $\pm$ 5.7 \\
& Tent~\cite{Wang2021-ho}\paperhsd & 84.5 $\pm$ 4.7 & 74.2 $\pm$ 13.9 & 76.4 $\pm$ 8.1 & 67.1 $\pm$ 10.1 & 80.1 $\pm$ 9.6 & 76.5 $\pm$ 6.5 & 19.0 $\pm$ 5.2 \\
& TASD~\cite{liu2022single}\paperhsd & 87.1 $\pm$ 2.5 & 76.4 $\pm$ 6.1 & 82.5 $\pm$ 5.2 & 76.0 $\pm$ 6.6 & 83.2 $\pm$ 6.7 & 81.1 $\pm$ 4.8 & 15.1 $\pm$ 5.3 \\

\hline

\multirow{5}{*}{\textit{\rotatebox{90}{DG}}}
& M-ADA~\cite{qiao2020learning}\paperhsd & 86.2 $\pm$ 4.4 & 74.7 $\pm$ 9.1 & 80.9 $\pm$ 4.9 & 69.7 $\pm$ 12.2 & 79.5 $\pm$ 9.3 & 78.2 $\pm$ 6.3 & 19.2 $\pm$ 7.1 \\
& BigAug~\cite{zhang2020generalizing}\paperhsd & 84.2 $\pm$ 5.0 & 73.9 $\pm$ 14.1 & 73.3 $\pm$ 7.7 & 74.7 $\pm$ 9.7 & 79.0 $\pm$ 6.8 & 77.0 $\pm$ 4.6 & 19.9 $\pm$ 4.7 \\
& HSD (CNN)~\cite{yi2024hallucinated}\paperhsd & 85.7 $\pm$ 3.2 & 78.9 $\pm$ 5.6 & 84.5 $\pm$ 3.0 & 78.6 $\pm$ 5.1 & 82.3 $\pm$ 4.1 & 82.0 $\pm$ 3.2 & 8.2 $\pm$ 2.4 \\
& HSD (Transformer)~\cite{yi2024hallucinated}\paperhsd & \underline{88.7} $\pm$ 2.2 & \underline{85.0} $\pm$ 2.5 & \underline{87.7} $\pm$ 1.9 & \textbf{87.2} $\pm$ 1.7 & \underline{90.3} $\pm$ 2.1 & \underline{87.8} $\pm$ 2.0 & \textbf{3.3} $\pm$ 2.5 \\

\cline{2-9}

& \textbf{SRCSM} & \textbf{90.8} $\pm$ 2.1 & \textbf{89.7} $\pm$ 3.5 & \textbf{90.5} $\pm$ 3.4 & \underline{86.2} $\pm$ 4.2 & \textbf{90.8} $\pm$ 2.3 & \textbf{89.6} $\pm$ 2.0 & \underline{3.4} $\pm$ 1.1 \\

\hline

\end{tabular}
\caption
[
Comparison to related work on prostate data in the HSD setup.
]
{
Comparison to related work on prostate data in the HSD setup.
Scores represent the mean and standard deviation of the DSC in \% when trained on prostate A and tested on the remaining target domains B, C, D, E and F individually as indicated in the respective column.
We also provide the mean and standard deviation of the HD95 in mm.
'Mean over Domains' refers to scores computed from the mean results of the individual domains.
Scores for related work are obtained from: \cite{yi2024hallucinated}~(\paperhsd).
Related work scores for HD95 were converted from pixel to mm by multiplication with the in-plane physical resolution of $0.4688 \text{ mm}$ of the resampled data.
Number of images in the test sets: B:~30, C:~19, D:~13, E:~12, F:~12.
}

\label{tab:iam_related_work_prostate_hsd_setup}
\end{table*}

%% file: tables/thale.t8.tex
\begin{table*}[t]
\centering
\footnotesize
\begin{tabular}{| l | c c c c c c | c |}
\hline

\multicolumn{1}{|c}{\cellcolor{lightgray} \textbf{Prostate}} & \multicolumn{7}{c|}{\cellcolor{lightgray} $\doubleunderscore$ $\rightarrow$ \textbf{Rest}} \\

\hline

\multicolumn{1}{|c|}{\multirow{3}{*}{\textbf{Method}}} & \multicolumn{7}{c|}{\multirow{1}{*}{DSC (\%) $\uparrow$}} \\

\cline{2-8}
& \multirow{2}{*}{A} & \multirow{2}{*}{B} & \multirow{2}{*}{C} & \multirow{2}{*}{D} & \multirow{2}{*}{E} & \multirow{2}{*}{F} & Mean over \\
& & & & & & & Domains \\

\hline

MaxStyle~\cite{chen2022maxstyle}\paperdesam & 81.25 & 70.27 & 62.09 & 58.18 & 70.04 & 67.77 & 68.27 $\pm$ 7.95 \\
CSDG~\cite{Ouyang2023-xf}\paperdesam & 80.72 & 68.00 & 59.78 & 72.40 & 68.67 & 70.78 & 70.06 $\pm$ 6.80 \\

MedSAM~\cite{ma2024segment}\paperdesam & 72.32 & 73.31 & 61.53 & 64.46 & 68.89 & 61.39 & 66.98 $\pm$ 5.28 \\
SAMed~\cite{zhang2023customized}\paperdesam & 73.61 & 75.89 & 58.61 & 73.91 & 66.52 & 72.85 & 70.23 $\pm$ 6.52 \\
DeSAM-B~\cite{gao2024desam}\paperdesam & 82.30 & 78.06 & \underline{66.65} & 82.87 & 77.58 & 79.05 & 77.75 $\pm$ 5.86 \\
DeSAM-P~\cite{gao2024desam}\paperdesam & \underline{82.80} & \underline{80.61} & 64.77 & \underline{83.41} & \underline{80.36} & \underline{82.17} & \underline{79.02} $\pm$ 7.08 \\

\hline

\textbf{SRCSM} & \textbf{89.76} $\pm$ 3.42 & \textbf{89.31} $\pm$ 5.37 & \textbf{84.31} $\pm$ 15.30 & \textbf{90.28} $\pm$ 3.07 & \textbf{88.41} $\pm$ 4.22 & \textbf{86.85} $\pm$ 9.69 & \textbf{88.15} $\pm$ 2.24 \\

\hline

\end{tabular}
\caption
[
Comparison to related work on prostate data in the DeSAM setup.
]
{
Comparison to related work on prostate data in the DeSAM setup.
Scores represent the mean and standard deviation of the DSC in \% when training on the source domain indicated in the respective column and tested on the union of all other target domains. 
For example, C $\rightarrow$ Rest refers to training on prostate C and averaging scores when testing on A, B, D, E and F.
'Mean over Domains' refers to scores computed from the mean results of the individual domains.
Differently to related work, we also provide standard deviations for our method.
Scores for related work are obtained from: \cite{gao2024desam}~(\paperdesam).
Number of images in the test sets: A~$\rightarrow$~Rest:~86, B~$\rightarrow$~Rest:~86, C~$\rightarrow$~Rest:~97, D~$\rightarrow$~Rest:~103, E~$\rightarrow$~Rest:~104, F~$\rightarrow$~Rest:~104.
}

\label{tab:iam_related_work_prostate_desam_setup}
\end{table*}

%% file: figures/thale6.tex
\begin{figure*}[t]
\centering
    \begin{subfigure}[t]{0.5\textwidth}
        \centering
        \includegraphics[width=\linewidth]{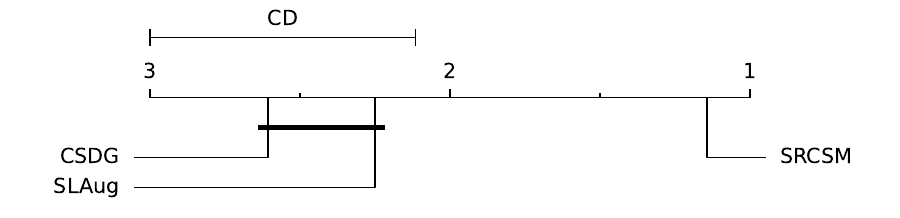}
        \caption{DSC}
    \end{subfigure}%
    ~ 
    \begin{subfigure}[t]{0.5\textwidth}
        \centering
        \includegraphics[width=\linewidth]{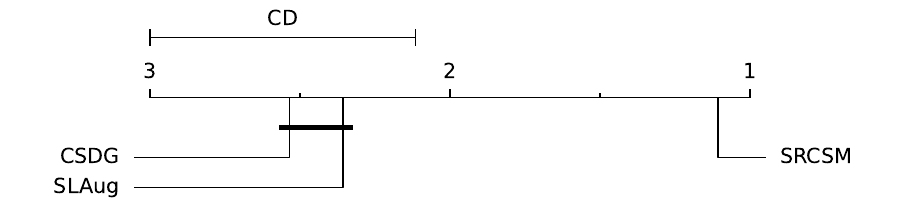}
        \caption{ASSD}
    \end{subfigure}
\caption{
The outcome of statistical analysis of multiple dataset generalization derived from 14 experimental setups reported in Tables~\ref{tab:abdominal_detailed} and~\ref{tab:related_work_mmwhs_7labels}.
Based on the Autorank package~\cite{Herbold2020-ea}, we compare three algorithms: SRCSM, SLAug and CSDG against each other with the non-parametric Friedman test and the post-hoc Nemenyi test.
Ranking based on (a)~DSC and (b)~ASSD performance metrics indicates superior dataset generalization capability of SRCSM.
Groups of algorithms that are not significantly different (at $p = 0.05$), i.e., within the critical difference (CD)~\cite{Demsar2006-zu} of the ranking, are connected.
}
\label{fig:ranking}
\end{figure*}

%% file: tables/thale.t9.tex
\begin{table*}[t]
\centering
\footnotesize
\resizebox{\textwidth}{!}{%
\begin{tabular}{| l | cc | cc | cc | cc | cc | cc cc cc | c |}
\hline

\multicolumn{1}{|c|}{\cellcolor{lightgray}} & \multicolumn{8}{c|}{\cellcolor{lightgray} \textbf{Abdominal}} & \multicolumn{2}{c|}{\cellcolor{lightgray} \textbf{Whole-Heart}} & \multicolumn{6}{c|}{\cellcolor{lightgray} \textbf{Prostate}} & \multicolumn{1}{c|}{\cellcolor{lightgray}} \\

\multicolumn{1}{|r|}{\cellcolor{lightgray} \textit{Source:}} & \multicolumn{1}{c}{\cellcolor{lightgray} \textbf{MR-1}} & \multicolumn{1}{c}{\cellcolor{lightgray} \textbf{MR-2}} &  \multicolumn{1}{c}{\cellcolor{lightgray} \textbf{CT-1}} & \multicolumn{1}{c}{\cellcolor{lightgray} \textbf{CT-2}} & \multicolumn{1}{c}{\cellcolor{lightgray} \textbf{MR-1}} & \multicolumn{1}{c}{\cellcolor{lightgray} \textbf{MR-2}} & \multicolumn{1}{c}{\cellcolor{lightgray} \textbf{CT-1}} & \multicolumn{1}{c|}{\cellcolor{lightgray} \textbf{CT-2}} & \multicolumn{1}{c}{\cellcolor{lightgray} \textbf{MR}} & \multicolumn{1}{c|}{\cellcolor{lightgray} \textbf{CT}} & \multicolumn{1}{c}{\cellcolor{lightgray} \textbf{Each}} & \multicolumn{1}{c}{\cellcolor{lightgray} \textbf{Each}} & \multicolumn{1}{c}{\cellcolor{lightgray} \textbf{Each}} & \multicolumn{1}{c}{\cellcolor{lightgray} \textbf{Each}} & \multicolumn{1}{c}{\cellcolor{lightgray} \textbf{Each}} & \multicolumn{1}{c|}{\cellcolor{lightgray} \textbf{Each}} & \multicolumn{1}{c|}{\cellcolor{lightgray} \textbf{Mean over}} \\

\multicolumn{1}{|r|}{\cellcolor{lightgray} \textit{Target:}} & \multicolumn{1}{c}{\cellcolor{lightgray} \textbf{CT-1}} & \multicolumn{1}{c}{\cellcolor{lightgray} \textbf{CT-1}} & \multicolumn{1}{c}{\cellcolor{lightgray} \textbf{MR-1}} & \multicolumn{1}{c}{\cellcolor{lightgray} \textbf{MR-1}} & \multicolumn{1}{c}{\cellcolor{lightgray} \textbf{CT-2}} & \multicolumn{1}{c}{\cellcolor{lightgray} \textbf{CT-2}} & \multicolumn{1}{c}{\cellcolor{lightgray} \textbf{MR-2}} & \multicolumn{1}{c|}{\cellcolor{lightgray} \textbf{MR-2}} & \multicolumn{1}{c}{\cellcolor{lightgray} \textbf{CT}} & \multicolumn{1}{c|}{\cellcolor{lightgray} \textbf{MR}} & \multicolumn{1}{c}{\cellcolor{lightgray} \textbf{A}} & \multicolumn{1}{c}{\cellcolor{lightgray} \textbf{B}} & \multicolumn{1}{c}{\cellcolor{lightgray} \textbf{C}} & \multicolumn{1}{c}{\cellcolor{lightgray} \textbf{D}} & \multicolumn{1}{c}{\cellcolor{lightgray} \textbf{E}} & \multicolumn{1}{c|}{\cellcolor{lightgray} \textbf{F}} & \multicolumn{1}{c|}{\cellcolor{lightgray} \textbf{Domains}} \\

\hline

\multirow{2}{*}{In-Domain}
& 92.1 & 92.1 & 91.2 & 91.2 & 96.1 & 96.1 & 96.5 & 96.5 & 91.5 & 85.6 & 90.8 & 93.0 & 93.7 & 90.2 & 89.7 & 90.8 & 92.3 \\
& $\pm$ 2.5 & $\pm$ 2.5 & $\pm$ 2.2 & $\pm$ 2.2 & $\pm$ 1.2 & $\pm$ 1.2 & $\pm$ 1.0 & $\pm$ 1.0 & $\pm$ 6.3 & $\pm$ 7.9 & $\pm$ 3.1 & $\pm$ 1.8 & $\pm$ 1.5 & $\pm$ 0.5 & $\pm$ 1.1 & $\pm$ 0.7 & $\pm$ 2.9 \\

\hline

\multirow{2}{*}{SRCSM}
& 89.8 & 91.6 & 88.7 & 90.0 & 86.6 & 93.4 & 93.7 & 94.5 & 87.4 & 85.7 & 91.2 & 89.6 & 81.7 & 90.2 & 89.7 & 89.0 & 89.6 \\
& $\pm$ 4.0 & $\pm$ 4.8 & $\pm$ 3.1 & $\pm$ 3.4 & $\pm$ 8.0 & $\pm$ 3.8 & $\pm$ 1.8 & $\pm$ 1.1 & $\pm$ 7.2 & $\pm$ 8.3 & $\pm$ 2.9 & $\pm$ 3.1 & $\pm$ 11.3 & $\pm$ 3.2 & $\pm$ 0.9 & $\pm$ 3.0 & $\pm$ 3.2 \\

\hline

\textit{Domain Gap} & \textit{2.3} & \textit{0.5} & \textit{2.5} & \textit{1.2} & \textit{9.5} & \textit{2.7} & \textit{2.8} & \textit{2.0} & \textit{4.0} & \textit{-0.1} & \textit{-0.4} & \textit{3.4} & \textit{12.0} & \textit{0.0} & \textit{0.0} & \textit{1.8} & \textit{2.8} \\

\hline
\end{tabular}
}
\caption
[
Comparison to in-domain baselines on abdominal, whole-heart and prostate domains.
]
{
Comparison to in-domain baselines on abdominal, whole-heart and prostate domains.
Scores represent the mean and standard deviation of the DSC in \%.
For the In-Domain baseline, scores were obtained by training on the target domain and evaluating on a held-out test set of that target domain.
Scores for SRCSM were obtained by training on the indicated source domain and evaluating on the same held-out test set.
Prostate scores for SRCSM represent the average over scores obtained when training on one domain and evaluating on the indicated target domain. 
For example, Each~$\rightarrow$~C refers to the average score of five settings, namely: A~$\rightarrow$~C, B~$\rightarrow$~C, D~$\rightarrow$~C, E~$\rightarrow$~C and F~$\rightarrow$~C.
Domain Gap refers to the remaining gap in performance between the in-domain baseline and SRCSM, which is computed as their difference.
For prostate domains, the standard deviation was computed from the mean scores achieved on the respective source domains.
The standard deviation shown in the column indicated with 'Mean over Domains' was computed over all dataset mean scores of the respective row.
Number of images in the test sets:
Abdominal: CT-1:~6, MR-1:~4, CT-2:~98, MR-2:~20.
Whole-Heart: CT:~40, MR:~40.
Prostate: A:~3, B:~3, C:~2, D:~2, E:~2, F:~2.
}
\label{tab:in_domain_comparison}
\end{table*}

%% file: figures/thale7.tex
\begin{figure*}[t]
\includegraphics[width=\textwidth]{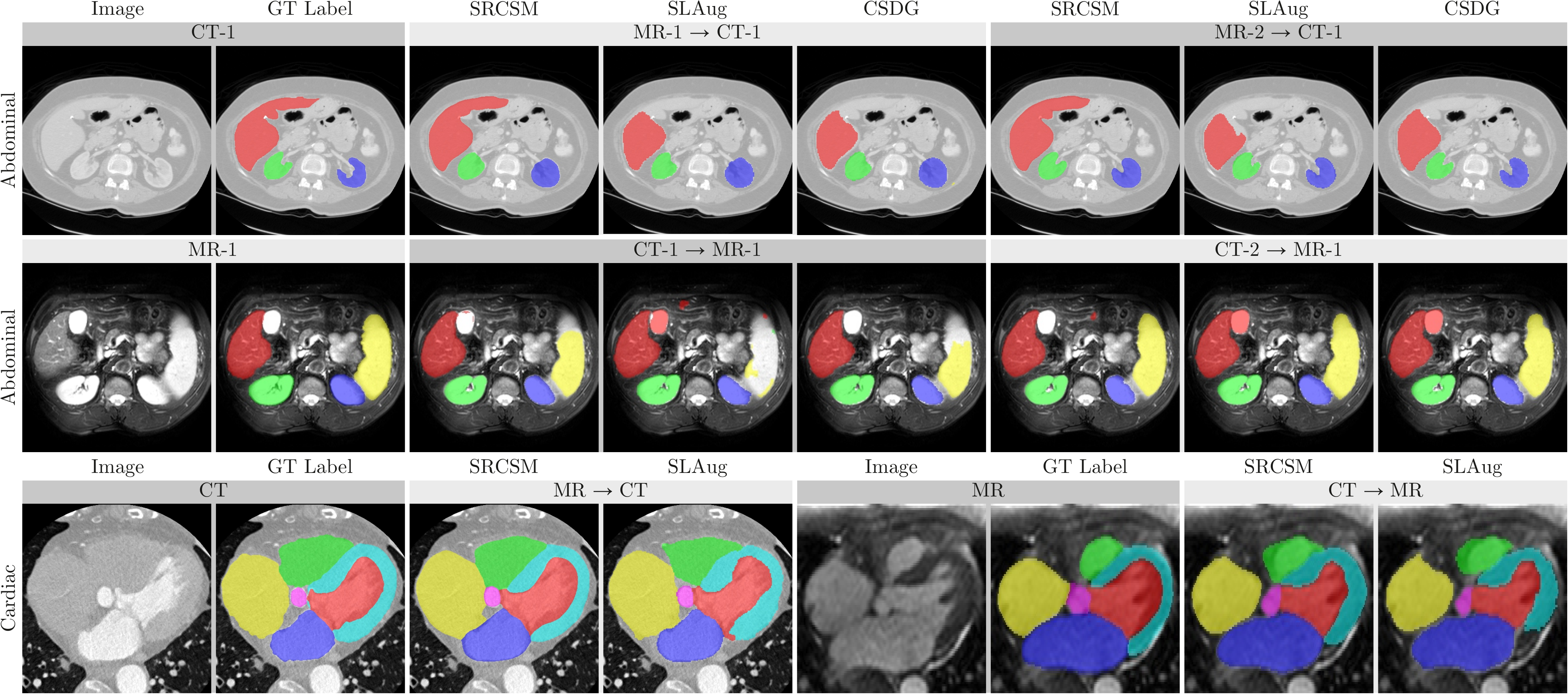}
\caption{
Qualitative results of our SRCSM method compared to the ground truth (GT) label, SLAug and CSDG.
Abdominal images are shown for CT-1 (row 1) when training on MR-1 (cols 3-5) and MR-2 (cols 6-8).
Results for MR-1 (row 2) are shown when training on CT-1 (cols 3-5) and CT-2 (cols 6-8).
Further, we provide cardiac whole-heart results (row 3) when adapting from MR to CT (cols 1-4) and from CT to MR (cols 5-8).
Column headers are provided above row 1 and row 3, respectively.
}
\label{fig:qualitative}
\end{figure*}

%% file: tables/thale.t10.tex
\begin{table*}
\centering
\footnotesize
\resizebox{\textwidth}{!}{%
\begin{tabular}{| l | l | c c | c c | c c | c c |}

\hline

\multicolumn{10}{|c|}{\cellcolor{lightgray} \textbf{Cardiac Whole-Heart $\rightarrow$ Cardiac Cine MR}} \\

\hline

\multicolumn{1}{|c}{\multirow{3}{*}{\textbf{Method}}} & & \multicolumn{2}{c|}{\textbf{MR $\rightarrow$ Cine MR (GE)}} & \multicolumn{2}{c|}{\textbf{MR $\rightarrow$ Cine MR (Phil.)}} & \multicolumn{2}{c|}{\textbf{MR $\rightarrow$ Cine MR (Siem.)}} & \multicolumn{2}{c|}{\textbf{MR $\rightarrow$ Cine MR (All)}} \\

\multicolumn{1}{|c}{\multirow{1}{*}{}} & & DSC & ASSD & DSC & ASSD & DSC & ASSD & DSC & ASSD \\
\multicolumn{1}{|c}{\multirow{1}{*}{}} & & (\%) $\uparrow$ & (mm) $\downarrow$ & (\%) $\uparrow$ & (mm) $\downarrow$ & (\%) $\uparrow$ & (mm) $\downarrow$ & (\%) $\uparrow$ & (mm) $\downarrow$ \\

\hline

\multicolumn{1}{|l|}{In-Domain$^{\ddagger}$} & \multirow{4}{*}{\textit{\rotatebox{90}{Diastolic}}} & -- & -- & -- & -- & -- & -- & 93.4 & -- \\
\multicolumn{1}{|l|}{In-Domain (ours)} & & 88.8 $\pm$ 6.2 & 1.0 $\pm$ 0.7 & 88.8 $\pm$ 3.8 & 0.7 $\pm$ 0.5 & 90.8 $\pm$ 3.0 & 0.7 $\pm$ 0.5 & 91.0 $\pm$ 3.1 & 0.7 $\pm$ 0.5 \\

\cdashline{1-1}[1pt/1pt]
\cdashline{3-10}[1pt/1pt]

\multicolumn{1}{|l|}{CDA} & & 79.0 $\pm$ 10.4 & 2.8 $\pm$ 4.3 & 80.0 $\pm$ 10.0 & 2.6 $\pm$ 4.0 & 74.0 $\pm$ 9.2 & 3.4 $\pm$ 4.7 & 75.0 $\pm$ 9.7 & 3.3 $\pm$ 4.6 \\
\multicolumn{1}{|l|}{SRCSM} & & 81.2 $\pm$ 3.0 & 1.8 $\pm$ 0.6 & 81.3 $\pm$ 3.2 & 1.6 $\pm$ 0.5 & 79.2 $\pm$ 4.1 & 1.9 $\pm$ 0.7 & 79.6 $\pm$ 4.0 & 1.8 $\pm$ 0.7 \\

\hline

\multicolumn{1}{|l|}{In-Domain$^{\ddagger}$} & \multirow{4}{*}{\textit{\rotatebox{90}{Systolic}}} & -- & -- & -- & -- & -- & -- & 91.0 & -- \\
\multicolumn{1}{|l|}{In-Domain (ours)} & & 89.1 $\pm$ 3.3 & 0.9 $\pm$ 0.4 & 88.9 $\pm$ 3.5 & 0.6 $\pm$ 0.4 & 89.5 $\pm$ 3.8 & 0.8 $\pm$ 0.6 & 89.9 $\pm$ 3.8 & 0.7 $\pm$ 0.5 \\

\cdashline{1-1}[1pt/1pt]
\cdashline{3-10}[1pt/1pt]

\multicolumn{1}{|l|}{CDA} & & 73.7 $\pm$ 11.6 & 4.7 $\pm$ 6.8 & 76.5 $\pm$ 10.0 & 3.5 $\pm$ 4.7 & 64.1 $\pm$ 16.5 & 7.1 $\pm$ 9.5 & 66.1 $\pm$ 16.2 & 6.6 $\pm$ 9.0 \\
\multicolumn{1}{|l|}{SRCSM} & & 79.0 $\pm$ 5.7 & 2.2 $\pm$ 1.0 & 79.9 $\pm$ 6.1 & 2.2 $\pm$ 1.3 & 75.3 $\pm$ 6.6 & 2.5 $\pm$ 1.0 & 76.0 $\pm$ 6.6 & 2.5 $\pm$ 1.0 \\

\hline
\hline

\multicolumn{1}{|c}{\multirow{3}{*}{\textbf{Method}}} & \multicolumn{1}{c|}{\multirow{1}{*}{}} & \multicolumn{2}{c|}{\textbf{CT $\rightarrow$ Cine MR (GE)}} & \multicolumn{2}{c|}{\textbf{CT $\rightarrow$ Cine MR (Phil.)}} & \multicolumn{2}{c|}{\textbf{CT $\rightarrow$ Cine MR (Siem.)}} & \multicolumn{2}{c|}{\textbf{CT $\rightarrow$ Cine MR (All)}} \\

\multicolumn{1}{|c}{\multirow{1}{*}{}} & & DSC & ASSD & DSC & ASSD & DSC & ASSD & DSC & ASSD \\
\multicolumn{1}{|c}{\multirow{1}{*}{}} & & (\%) $\uparrow$ & (mm) $\downarrow$ & (\%) $\uparrow$ & (mm) $\downarrow$ & (\%) $\uparrow$ & (mm) $\downarrow$ & (\%) $\uparrow$ & (mm) $\downarrow$ \\

\hline

\multicolumn{1}{|l|}{In-Domain$^{\ddagger}$} & \multirow{4}{*}{\textit{\rotatebox{90}{Diastolic}}} & -- & -- & -- & -- & -- & -- & 93.4 & -- \\
\multicolumn{1}{|l|}{In-Domain (ours)} & & 88.8 $\pm$ 6.2 & 1.0 $\pm$ 0.7 & 88.8 $\pm$ 3.8 & 0.7 $\pm$ 0.5 & 90.8 $\pm$ 3.0 & 0.7 $\pm$ 0.5 & 91.0 $\pm$ 3.1 & 0.7 $\pm$ 0.5 \\

\cdashline{1-1}[1pt/1pt]
\cdashline{3-10}[1pt/1pt]

\multicolumn{1}{|l|}{CDA} & & 17.3 $\pm$ 11.4 & 31.0 $\pm$ 19.4 & 17.2 $\pm$ 8.3 & 28.4 $\pm$ 14.3 & 15.5 $\pm$ 9.3 & 38.2 $\pm$ 22.3 & 15.8 $\pm$ 9.5 & 36.7 $\pm$ 21.7 \\
\multicolumn{1}{|l|}{SRCSM} & & 77.7 $\pm$ 10.2 & 2.7 $\pm$ 3.7 & 79.7 $\pm$ 5.5 & 2.4 $\pm$ 4.0 & 76.7 $\pm$ 5.1 & 2.1 $\pm$ 0.8 & 77.0 $\pm$ 6.0 & 2.2 $\pm$ 1.8 \\

\hline

\multicolumn{1}{|l|}{In-Domain$^{\ddagger}$} & \multirow{4}{*}{\textit{\rotatebox{90}{Systolic}}} & -- & -- & -- & -- & -- & -- & 91.0 & -- \\
\multicolumn{1}{|l|}{In-Domain (ours)} & & 89.1 $\pm$ 3.3 & 0.9 $\pm$ 0.4 & 88.9 $\pm$ 3.5 & 0.6 $\pm$ 0.4 & 89.5 $\pm$ 3.8 & 0.8 $\pm$ 0.6 & 89.9 $\pm$ 3.8 & 0.7 $\pm$ 0.5 \\

\cdashline{1-1}[1pt/1pt]
\cdashline{3-10}[1pt/1pt]

\multicolumn{1}{|l|}{CDA} & & 24.1 $\pm$ 16.9 & 26.3 $\pm$ 21.8 & 26.1 $\pm$ 16.2 & 25.8 $\pm$ 20.9 & 13.5 $\pm$ 10.7 & 41.1 $\pm$ 26.1 & 15.6 $\pm$ 12.8 & 38.3 $\pm$ 25.9 \\
\multicolumn{1}{|l|}{SRCSM} & & 76.4 $\pm$ 3.8 & 2.7 $\pm$ 1.1 & 74.3 $\pm$ 8.5 & 4.4 $\pm$ 6.0 & 73.7 $\pm$ 6.5 & 3.0 $\pm$ 1.9 & 74.0 $\pm$ 6.4 & 3.1 $\pm$ 2.3 \\

\hline

\end{tabular}
}
\caption{
Quantitative results when training on whole-heart and testing on cardiac cine MR data.
We evaluate the performance of our SRCSM method to a method that only relies on conventional data augmentation (CDA), as well as to two In-Domain baselines.
As source domain we used whole-heart MR (top half of rows) and CT (bottom half of rows), respectively.
For evaluation, we separated the cine dataset into diastolic and systolic scans, as well as by scanner manufacturer into General Electric (GE), Philips (Phil.) and Siemens (Siem.).
To allow a comparison to In-Domain related work in literature, we also show results without separating by scanner manufacturer (All).
Presented are the mean and standard deviation of the DSC in \% as well as the ASSD in mm.
The In-Domain scores indicated by~$^{\ddagger}$ are based on~\cite{fulton2021deformable} as obtained from~\cite{martin2023deep}.
Number of images in the test sets: GE:~19, Phil.:~10, Siem.:~128, All:~157.
}
\label{tab:whs_to_cine}
\end{table*}

%% file: tables/thale.t11.tex
\begin{table*}
\centering
\footnotesize

\resizebox{\textwidth}{!}{%
\begin{tabular}{|l| c c | c c | c c | c c |}

\hline

\multicolumn{1}{|c}{\multirow{1}{*}{\cellcolor{lightgray}}} & \multicolumn{4}{c|}{\cellcolor{lightgray} \textbf{Abdominal}} & \multicolumn{4}{c|}{\cellcolor{lightgray} \textbf{Cardiac:} 7 Labels} \\

\hline

\multicolumn{1}{|c|}{\multirow{3}{*}{\textbf{Method}}} & \multicolumn{2}{c|}{\textbf{CT-1 $\rightarrow$ MR-1}} & \multicolumn{2}{c|}{\textbf{MR-1 $\rightarrow$ CT-1}} & \multicolumn{2}{c|}{\textbf{CT $\rightarrow$ MR}} & \multicolumn{2}{c|}{\textbf{MR $\rightarrow$ CT}} \\

\multicolumn{1}{|c|}{} & DSC & ASSD & DSC & ASSD & DSC & ASSD & DSC & ASSD \\

& (\%) $\uparrow$ & (mm) $\downarrow$ & (\%) $\uparrow$ & (mm) $\downarrow$ & (\%) $\uparrow$ & (mm) $\downarrow$ & (\%) $\uparrow$ & (mm) $\downarrow$ \\

\hline

CDA & 8.3 $\pm$ 7.5 & 40.4 $\pm$ 9.7 & 5.6 $\pm$ 3.7 & 111.1 $\pm$ 11.5 & 34.5 $\pm$ 10.5 & 13.6 $\pm$ 3.7  & 36.6 $\pm$ 13.5 & 19.4 $\pm$ 7.0 \\
CDA+RC & 85.4 $\pm$ 7.8 & 3.1 $\pm$ 3.0 & 85.0 $\pm$ 9.0 & 4.5 $\pm$ 3.3 & 82.9 $\pm$ 3.6 & 2.2 $\pm$ 0.8  & 84.4 $\pm$ 3.2 & 2.0 $\pm$ 0.6 \\
CDA+SRC$_{\text{binary}}$ & 65.3 $\pm$ 16.8 & 8.8 $\pm$ 7.0 & 69.5 $\pm$ 18.8 & 8.6 $\pm$ 7.8 & 81.5 $\pm$ 5.3 & 2.3 $\pm$ 0.9  & 78.6 $\pm$ 7.0 & 3.2 $\pm$ 1.4 \\
CDA+SRC & \underline{88.1} $\pm$ 5.1 & \underline{2.3} $\pm$ 1.2 & 83.3 $\pm$ 9.6 & 4.9 $\pm$ 5.0 & 84.2 $\pm$ 3.5 & 2.1 $\pm$ 0.9  & 86.9 $\pm$ 2.3 & 1.7 $\pm$ 0.4 \\

\cdashline{1-9}[1pt/1pt]

CDA+SM & 58.0 $\pm$ 18.8 & 9.1 $\pm$ 7.6 & 51.1 $\pm$ 14.7 & 27.1 $\pm$ 18.5 & 68.1 $\pm$ 11.9 & 7.2 $\pm$ 3.4  & 68.1 $\pm$ 11.9 & 7.2 $\pm$ 3.4 \\
CDA+RC+SM & 85.1 $\pm$ 9.6 & 2.7 $\pm$ 2.7 & \underline{87.7} $\pm$ 8.8 & \underline{3.7} $\pm$ 3.2 & \underline{84.8} $\pm$ 3.0 & \underline{1.9} $\pm$ 0.6  & 87.1 $\pm$ 2.0 & 1.7 $\pm$ 0.4 \\
CDA+SRC$_{\text{binary}}$+SM & 87.9 $\pm$ 5.6 & \underline{2.3} $\pm$ 1.7 & \textbf{88.0} $\pm$ 9.3 & \textbf{3.6} $\pm$ 3.4 & 84.3 $\pm$ 3.5 & 2.0 $\pm$ 0.8  & \underline{87.7} $\pm$ 2.4 & \underline{1.6} $\pm$ 0.5 \\
\textbf{CDA+SRC+SM (SRCSM)} & \textbf{89.5} $\pm$ 3.0 & \textbf{1.8} $\pm$ 0.8 & 87.6 $\pm$ 8.2 & \underline{3.7} $\pm$ 3.1 & \textbf{85.7} $\pm$ 2.8 & \textbf{1.8} $\pm$ 0.6  & \textbf{88.7} $\pm$ 2.1 & \textbf{1.4} $\pm$ 0.4 \\

\hline

\end{tabular}
}
\caption{
Ablation study of our SRCSM method.
Presented are the mean and standard deviation of the DSC in \% as well as the ASSD in mm.
Scores for the whole-heart dataset refer to the setting using seven labels.
Abbreviations:
Conventional Data Augmentation (CDA),
Random Convolution (RC),
Semantic-aware Random Convolution (SRC) and
Source Matching (SM).
CDA is described in Section~\ref{sec:implementation_details}.
RC is a semantic-\textit{un}aware SRC variant which employs a single random convolution network to augment the whole image similar to CSDG~\cite{Ouyang2023-xf}.
SRC$_{\text{binary}}$ refers to the implementation of Eq.~\eqref{eq:sarc_simplified}.
SRC and SM are our contributions as respectively described in Section~\ref{sec:training_time_strategies} and~\ref{sec:test_time_strategies}.
Number of images in the test sets:
Abdominal: MR-1:~20, CT-1:~30.
Cardiac: MR:~20, CT:~20.
}
\label{tab:ablation}
\end{table*}

%% file: figures/thale8.tex
\begin{figure*}[!t]
\centering
\includegraphics[width=0.75\textwidth]{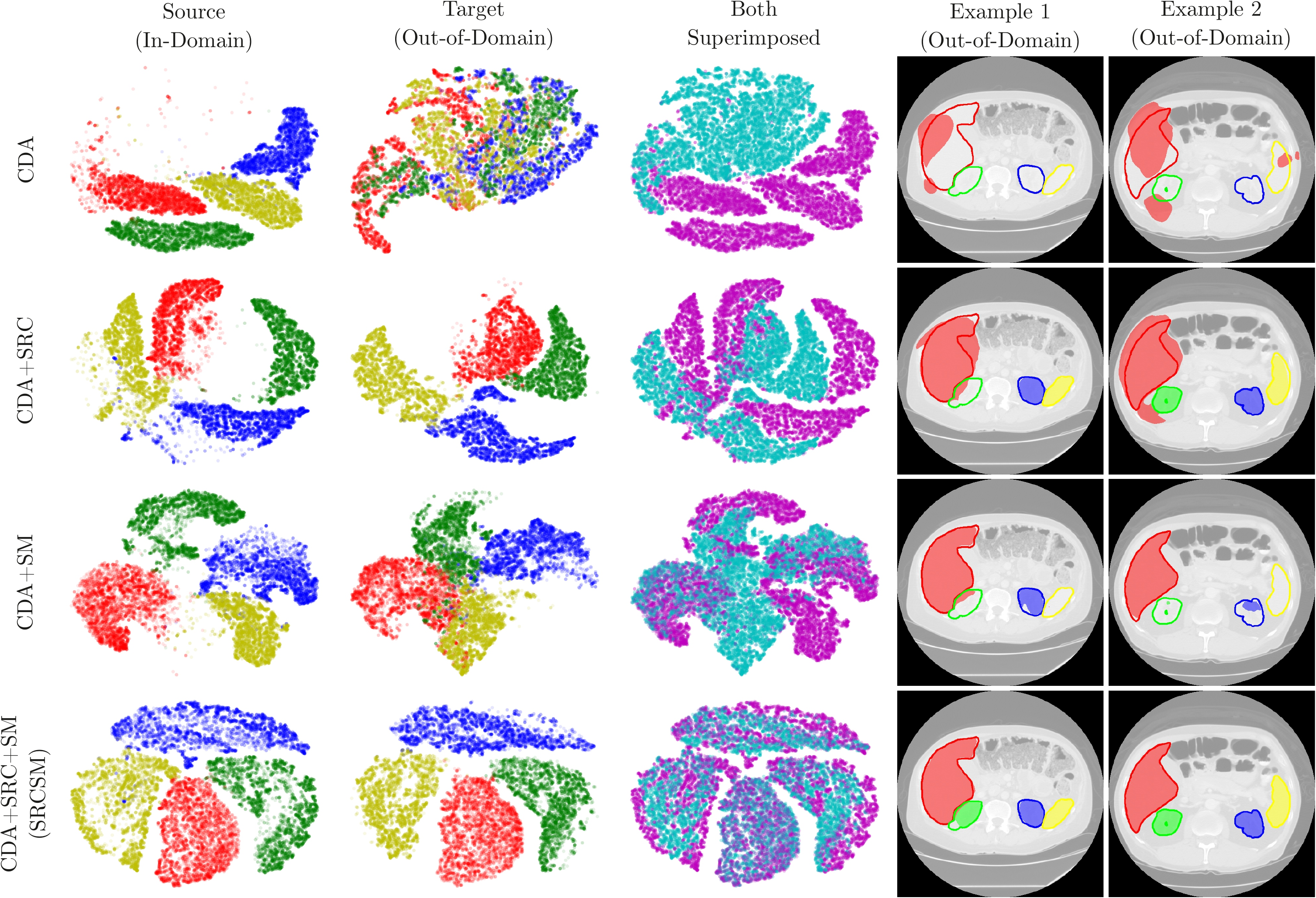}
\caption{
Intermediate feature representations after t-SNE~\cite{van2008visualizing} dimensionality reduction (col~1-3) and exemplary images (col~4-5) from different models used in our ablation study of the abdominal \gls{mr}-1~$\rightarrow$~\gls{ct}-1 experiment.
The plots correspond to the CDA baseline (row~1), CDA+SRC (row~2), CDA+SM (row~3) and the proposed method CDA+SRC+SM (i.e., SRCSM, row~4).
We provide separate plots for source points (in-domain, col~1), target points (out-of-domain, col~2) and both superimposed onto one another (col~3).
The colors in col~1-2 represent points labeled as liver (red), left kidney (green), right kidney (blue) and spleen (yellow).
The colors in col~3 represent source (magenta) and target points (cyan).
All points are plotted with transparency to convey point density.
In col~4-5, we provide exemplary out-of-domain images to complement the t-SNE plots.
The contour lines represent the ground truth segmentation per label, while the colored semi-transparent areas represent predictions produced by the same models used to generate the plots in the same row.
}
\label{fig:tsne_all}
\end{figure*}

%% file: figures/thale9.tex
\begin{figure*}[!t]
\centering
\includegraphics[width=0.75\textwidth]{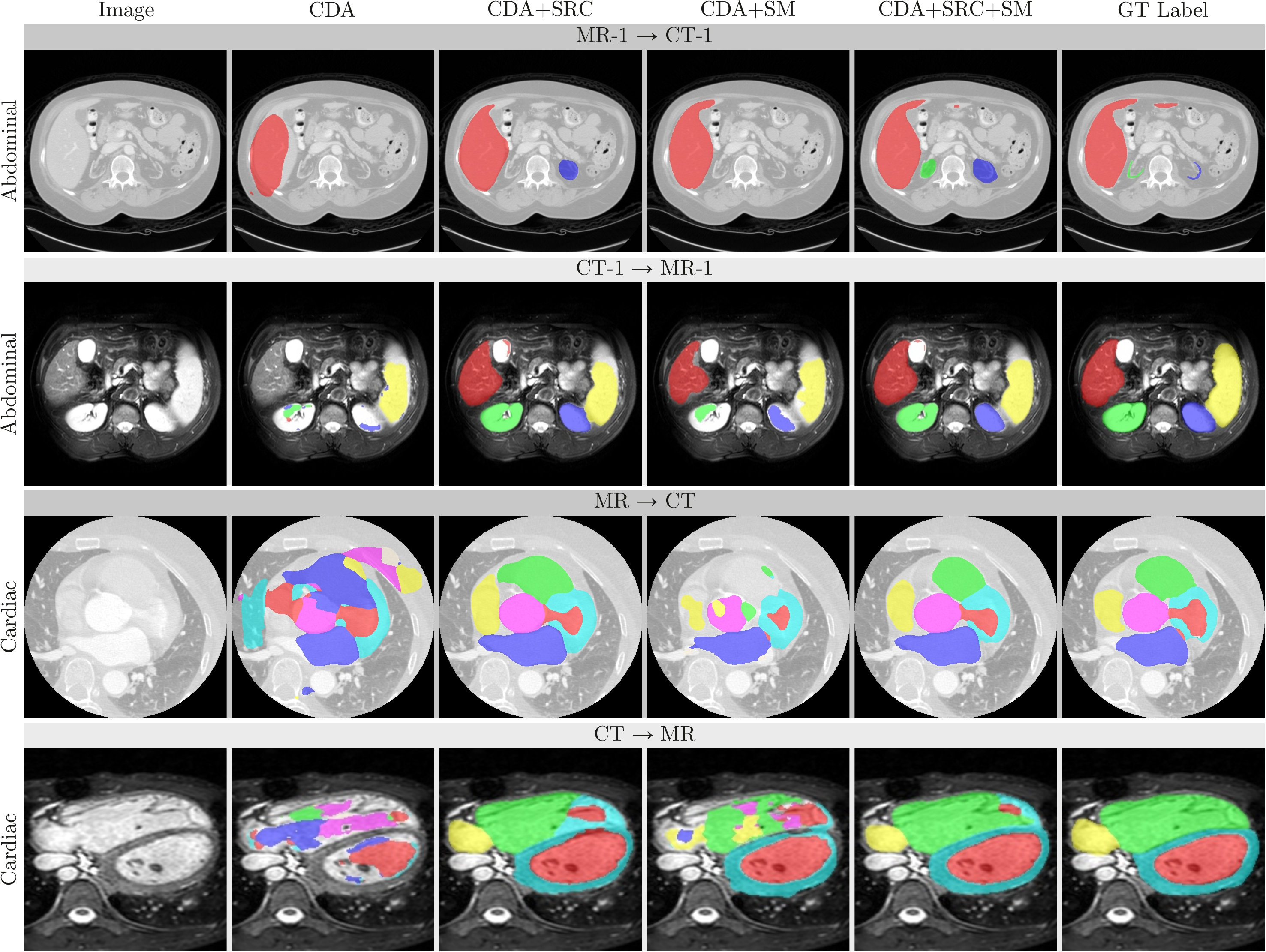}
\caption{
Qualitative error analysis with examples from the ablation study showing how our contributions affect model predictions.
Shown are the image (col~1), the image overlaid with the prediction of the CDA baseline (col~2), CDA+SRC (col~3), CDA+SM (col~4), the proposed SRCSM method which employs CDA+SRC+SM (col~5), as well as the image overlaid with the ground truth (col~6).
We present examples from the following settings:
abdominal MR-1~$\rightarrow$~CT-1 (row~1) and
CT-1~$\rightarrow$~MR-1 (row~2), as well as 
cardiac MR~$\rightarrow$~CT (row~3) and
CT~$\rightarrow$~MR (row~4) with 7 labels.
}
\label{fig:case_level_error_analysis}
\end{figure*}

%% file: our_paper/conclusion.tex
\section{Conclusion}

In this work we propose \gls{srcsm}, a single-source \gls{dg} method for semantic segmentation consisting of two contributions.
First, \gls{src} diversifies the source domain through semantic-aware random convolution during training.
Second, \gls{sm} efficiently aligns individual target domain images with the distribution of the source domain at test-time by employing the average histogram of the source dataset, which is computed and stored alongside the model after training finishes.
Therefore, \gls{srcsm} tackles the challenging scenario of having \textit{no} target domain data available during training, while also requiring \textit{no} access to source domain images at test-time and \textit{without} demanding any target specific optimization procedure at test-time.
Our comprehensive evaluation involves cross-modality and cross-center generalization and includes a wide range of comparison methods for abdominal organ, whole-heart and prostate segmentation in various settings.  
In addition, we also assess our \gls{srcsm} method when training on whole-heart \gls{ct} or whole-heart \gls{mr} data and testing on the diastolic and systolic phase of cine \gls{mr} data obtained from several scanner manufacturers.
\gls{srcsm} outperforms related work in most observed settings, often by a respectable margin, and even achieves results that match the in-domain performance in several settings.
With these results, our large-scale evaluation confirms the generality and wide applicability of the proposed method to various \gls{mr} and \gls{ct} domains, establishing \gls{srcsm} as the new state-of-the-art in domain generalization for medical image segmentation.